\documentclass[10pt,twocolumn,letterpaper]{article}

\usepackage{cvpr}              % To produce the CAMERA-READY version
\usepackage[dvipsnames]{xcolor}

\definecolor{LightCyan}{rgb}{0.88,1,1}
\definecolor{rowred}{rgb}{0.35,1,1}

\definecolor{rowlightblue}{rgb}{1,0.88,1}
\definecolor{rowblue}{rgb}{1,0.55,1}

\NewDocumentCommand{\rot}{O{45} O{1em} m}{\makebox[#2][l]{\rotatebox{#1}{#3}}}%

\definecolor{cvprblue}{rgb}{0.21,0.49,0.74}
\usepackage[pagebackref,breaklinks,colorlinks,allcolors=cvprblue]{hyperref}
\usepackage{color}
\usepackage{pifont}
\usepackage{multirow}
\usepackage{fix-cm} 
\def\paperID{12579} % *** Enter the Paper ID here
\def\confName{CVPR}
\def\confYear{2025}

\title{AG-VPReID: A Challenging Large-Scale Benchmark \\for Aerial-Ground Video-based Person Re-Identification }

\author{Huy Nguyen$^1$, \quad Kien Nguyen$^1$, \quad 
Akila Pemasiri$^1$, \quad Feng Liu$^2$, \\
Sridha Sridharan$^1$, \quad Clinton Fookes$^1$\\
\textsuperscript{1} School of Electrical Engineering and Robotics, Queensland University of Technology\\
\textsuperscript{2} Department of Computer Science, Drexel University\\
{\tt\small $^1$\{t497.nguyen, k.nguyenthanh, a.thondilege, s.sridharan, c.fookes\}@qut.edu.au,}\\
{\tt\small $^2$fl397@drexel.edu}
}

\begin{document}
\maketitle

\begin{abstract}
We introduce AG-VPReID, a new large-scale dataset for aerial-ground video-based person re-identification (ReID) that comprises 6,632 subjects, 32,321 tracklets and over 9.6 million frames captured by drones (altitudes ranging from 15–120m), CCTV, and wearable cameras. This dataset offers a real-world benchmark for evaluating the robustness to significant viewpoint changes, scale variations, and resolution differences in cross-platform aerial-ground settings. 
In addition, to address these challenges, we propose AG-VPReID-Net, an end-to-end framework composed of three complementary streams: (1) an Adapted Temporal-Spatial Stream addressing motion pattern inconsistencies and facilitating temporal feature learning,
(2) a Normalized Appearance Stream leveraging physics-informed techniques to tackle resolution and appearance changes, and (3) a Multi-Scale Attention Stream handling scale variations across drone altitudes. We integrate visual-semantic cues from all streams to form a robust, viewpoint-invariant whole-body representation. 
Extensive experiments demonstrate that AG-VPReID-Net outperforms state-of-the-art approaches on both our new dataset and existing video-based ReID benchmarks, showcasing its effectiveness and generalizability. 
Nevertheless, the performance gap observed on AG-VPReID across all methods underscores the dataset’s challenging nature.
The dataset, code and trained models are available at 
\href{https://github.com/agvpreid25/AG-VPReID-Net}{AG-VPReID-Net}.

\end{abstract}

\section{Introduction}
\label{sec:intro}
Video-based person re-identification (ReID) is a challenging and in-demand task, with significant real-world applications in surveillance, search and rescue operations, and urban monitoring \cite{liu2021watching,pan2023pose,ASAS}. 
While traditional ReID methods focus on ground-based cameras \cite{zhang2020multi,fu2019sta}, the integration of aerial perspectives through aerial-ground person ReID presents a paradigm shift in this field \cite{Nguyen2024AGReIDv2BA}. This approach enables the identification and matching of individuals across non-overlapping aerial and ground-based camera views, substantially enhancing situational awareness and response times in complex environments \cite{AGReIDv1,CARGO}. 
The motivation behind this research stems from the increasing deployment of aerial platforms, such as unmanned aerial vehicles (UAVs), which provide unique vantage points that complement ground-based observations. However, the development of robust aerial-ground ReID systems faces a significant challenge: the scarcity of diverse and large-scale datasets that capture the nuances of both aerial and ground perspectives.  As demonstrated by ImageNet \cite{deng2009imagenet}, large and diverse benchmarks are crucial for deep learning based methods, indicating a need for a comprehensive ReID dataset integrating multiple platforms, environments, and real-world challenges.

\begin{table*}[t] 
    \centering
    \scriptsize
    % \footnotesize
    \begin{tabular}{l|c|c|c|c|c|c|c|c|c|c|c|c}
    \toprule
    %\textbf{Dataset} & \textbf{Year} & \textbf{IDs} & \textbf{Trk} & \textbf{Fr (M)} & \textbf{CV} & \textbf{LT} & \textbf{Att} & \multicolumn{3}{c}{\textbf{Platform}} & \textbf{Dur } & \textbf{Alt (m)} \\
    \multirow{2}{*}{Dataset} & \multirow{2}{*}{Year} & \multirow{2}{*}{\#Identities} & \multirow{2}{*}{\#Tracklets} & \multirow{2}{*}{\#Frames (M)} & \multirow{2}{*}{\#CV}& \multirow{2}{*}{CC}& \multirow{2}{*}{Att.} &  \multicolumn{3}{c|}{Platform} & \multirow{2}{*}{Dur.} & \multirow{2}{*}{Altitude (m)} \\
    
    \cmidrule(lr){9-11}
    & & & & & & & & Ground & Wearable & Aerial & & \\
    \midrule
    \midrule
    MARS~\cite{mars} & 2016 & 1,261 & 20,478 & 1.19 & 6 & \textcolor{red}{\ding{55}} & \textcolor{red}{\ding{55}} & \textcolor{green}{\ding{51}} & \textcolor{red}{\ding{55}} & \textcolor{red}{\ding{55}} & - & - \\
    LSVID~\cite{GLTR_LS-VID} & 2019 & 3,772 & 14,943 & 2.98 & 15 & \textcolor{red}{\ding{55}} & \textcolor{red}{\ding{55}} & \textcolor{green}{\ding{51}} & \textcolor{red}{\ding{55}} & \textcolor{red}{\ding{55}} & 4 & -  \\
    VCCR~\cite{han20223d} & 2022 & 392 & 4,384 & 0.15 & 1 & \textcolor{green}{\ding{51}} & \textcolor{red}{\ding{55}} & \textcolor{green}{\ding{51}} & \textcolor{red}{\ding{55}} & \textcolor{red}{\ding{55}} & 90 & - \\
    CCVID~\cite{gu2022CAL} & 2022 & 226 & 2,856 & 0.34 & 1 & \textcolor{green}{\ding{51}} & \textcolor{red}{\ding{55}} & \textcolor{green}{\ding{51}} & \textcolor{red}{\ding{55}} & \textcolor{red}{\ding{55}} & - & -  \\
    MEVID~\cite{Davila2023mevid} & 2023 & 158 & 8,092 & \textbf{10.46} & \textbf{33} & \textcolor{green}{\ding{51}} & \textcolor{red}{\ding{55}} & \textcolor{green}{\ding{51}} & \textcolor{red}{\ding{55}} & \textcolor{red}{\ding{55}} & 73 & - \\
     \hline
    PDestre~\cite{kumar2020p} & 2020 & 253 & 1,894 & 0.10 & 1 & \textcolor{green}{\ding{51}} & \textcolor{green}{\ding{51}} & \textcolor{red}{\ding{55}} & \textcolor{red}{\ding{55}} & \textcolor{green}{\ding{51}} & - & 5-6  \\
    G2A-VReID~\cite{zhang2024cross} & 2024 & 2,788 & 5,576 & 0.18 & 2 & \textcolor{red}{\ding{55}} & \textcolor{red}{\ding{55}} & \textcolor{green}{\ding{51}} & \textcolor{red}{\ding{55}} & \textcolor{green}{\ding{51}} & - & 20 - 60 \\
    % \hline
    \textbf{AG-VPReID} & {2024} & \textbf{6,632} & \textbf{32,321} & 9.6 & 6 & \textcolor{green}{\ding{51}} & \textcolor{green}{\ding{51}} & \textcolor{green}{\ding{51}} & \textcolor{green}{\ding{51}} & \textcolor{green}{\ding{51}} & 20 & \textbf{15 - 120} \\
    \bottomrule
    \end{tabular}
    \vspace{-6px}
    \caption{\footnotesize Comparison of AG-VPReID with existing video-based person ReID datasets. Above: ground-based datasets, Below: aerial-based datasets. CV: Camera Views, CC: Clothes-Change, Att.: Attributes (Soft-biometrics annotations), Dur.: Duration (days).   }
    \label{tab:dataset_comparison}
\end{table*}

Initial efforts in aerial-ground person ReID have focused primarily on image-based tasks. For instance, Nguyen \emph{et al.}~\cite{AGReIDv1} pioneer this area by releasing the first aerial-ground ReID dataset, which includes images from one drone and one CCTV camera capturing 21,983 images of 388 identities.
They later expand the dataset to 100,502 images of 1,615 individuals \cite{Nguyen2024AGReIDv2BA}. Recently, Zhang \emph{et al.}~\cite{CARGO} collect a synthetic dataset named CARGO, containing 108,563 images representing 5,000 subjects, to complement real-world datasets. Within video-based tasks, Zhang \emph{et al.}~\cite{zhang2024cross} collect a video-based dataset called G2A-VReID, which consists of 185,907 images and 5,576 tracklets from one drone and one CCTV camera, featuring 2,788 identities.
{Building on these advancements, G2A-VReID could be expanded compared to ground-based datasets like MARS~\cite{mars}, which includes 20,000 tracklets and 1.19 million frames from six cameras.}
{While current aerial-ground datasets are valuable, increasing identity variation and environmental diversity would improve model robustness for real-world applications.}

In light of this, we introduce \textbf{AG-VPReID}, a comprehensive large-scale benchmark dataset for Aerial-Ground Video-based Person ReID. AG-VPReID comprises 6,632 subjects, 32,321 tracklets, and over 9.6 million frames, captured across multiple dates and times of day using a combination of three platforms: aerial drones operating at various altitudes (15-120m), stationary CCTV cameras and wearable mobile cameras. This dataset significantly surpasses existing video-based ReID benchmarks in terms of scale, diversity, and real-world applicability with the highest number of identities, the highest number of tracklets, the highest drone flying altitudes, and the most diverse platforms. The key characteristics of AG-VPReID include: drastic view changes between aerial and ground perspectives; a large number of annotated identities across multiple sessions; rich outdoor scenarios with varying environmental conditions; significant differences in resolution between aerial and ground footage; and both controlled scenarios with clothing changes and in-the-wild pedestrian traffic. 

Aerial-ground person ReID presents unique challenges due to significant appearance variations between aerial and ground-level views. These variations include extreme viewpoint differences, drastic changes in resolution and scale, partial occlusions, and temporal discontinuities caused by high-flying altitudes and long-range captures.
Traditional video-based person ReID methods, although effective in ground-based settings \cite{liu2023video,yu2024tf}, often struggle in aerial-ground scenarios due to the complex combination of inconsistent motion patterns and the aforementioned variations.

To address these challenges, we introduce \textbf{AG-VPReID-Net}, an end-to-end framework for Aerial-Ground Video-based Person Re-Identification. Unlike existing state-of-the-art methods focused on single-view or ground scenarios, AG-VPReID-Net features three complementary streams tailored for aerial-ground challenges:
\textbf{\emph{i)}} An Adapted Temporal-Spatial Stream enhances traditional temporal modeling by integrating identity-specific memory and temporal shape analysis. This improves the extraction of consistent motion patterns and body shape representations, addressing the temporal discontinuity and motion inconsistencies of aerial footage, outperforming standard LSTM \cite{eom2021video,hou2021bicnet} or 3D CNN \cite{li2019multi} approaches;
\textbf{\emph{ii)}} A Normalized Appearance Stream addresses the resolution and appearance differences between aerial and ground views by using UV maps aggregation across frames for a normalized appearance representation. This provides robustness against pose changes, viewpoint shifts, and varying image quality, excelling where current appearance-based methods \cite{liu2019spatial,li2019global} falter; and
% 
% %
\textbf{\emph{iii)}} A Multi-Scale Attention Stream addresses scale variations inherent in aerial-ground data by incorporating multi-scale feature extraction, motion analysis, temporal context, and a transformer decoder, effectively improving identification across drone altitudes compared to single-scale \cite{hou2020temporal,yan2020learning} methods.
By integrating these streams, AG-VPReID-Net offers incremental improvements in aerial-ground video-based re-identification, highlighting its potential in addressing this challenging scenario.

In summary, our main contributions are as follows: 

(1) We introduce AG-VPReID, a challenging large-scale benchmark for aerial-ground video-based person ReID, bridging the gap with a diverse dataset that captures nuanced challenges from both aerial and ground perspectives.

(2) We propose AG-VPReID-Net, an innovative end-to-end framework that integrates adapted temporal-spatial processing, normalized appearance representation, and multi-scale attention mechanisms to effectively address the challenges of aerial-ground ReID.

(3) AG-VPReID-Net sets new state-of-the-art performance on the AG-VPReID and existing video-based ReID benchmarks, demonstrating our approach’s effectiveness and generalizability across different settings.

\begin{figure*}[!htbp] 
    \centering
    \includegraphics[width=0.9\linewidth]{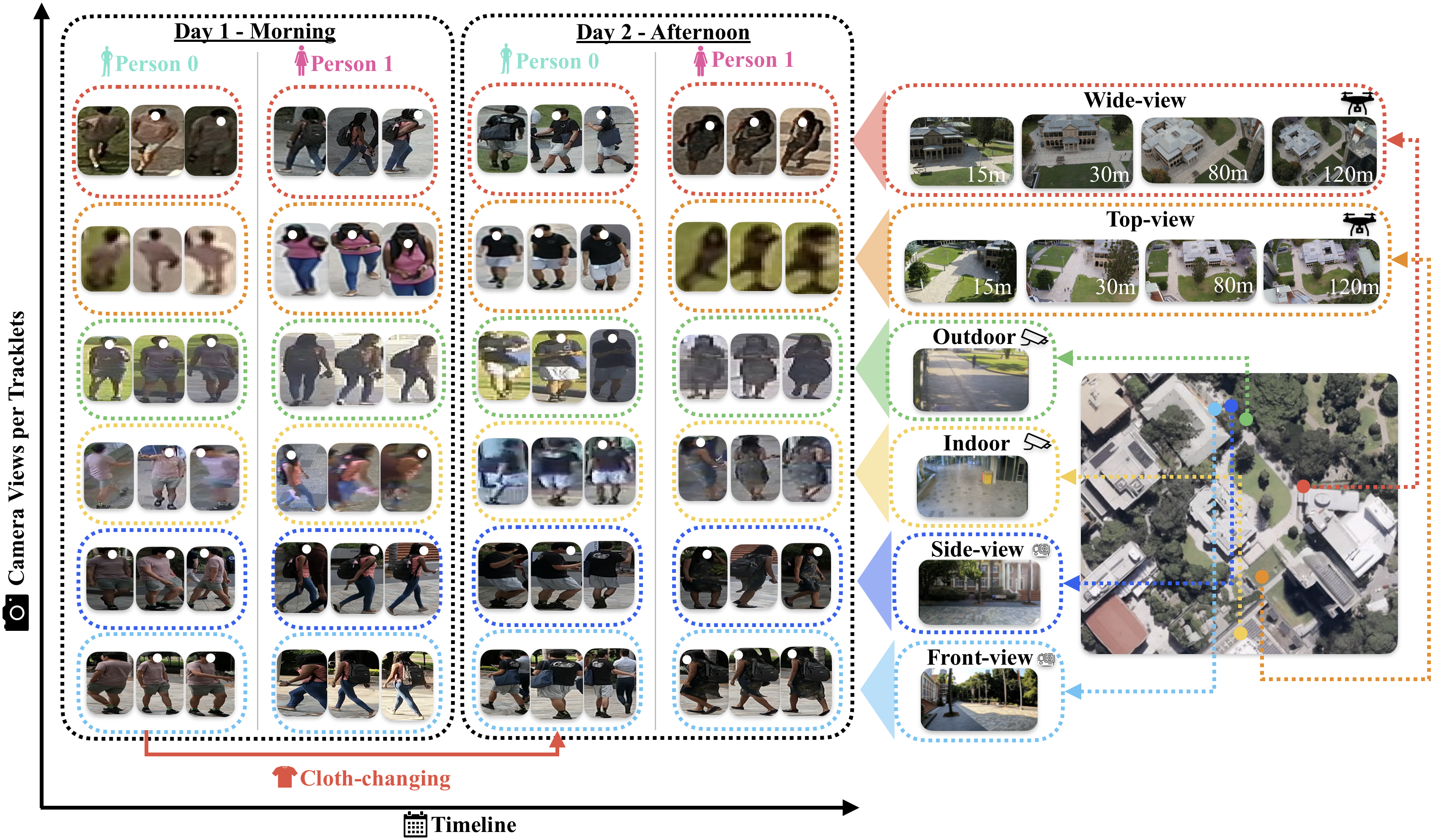}
    \caption{ Our AG-VPReID dataset was captured using a variety of six cameras, including aerial drones, CCTVs, and GoPros. Sample images and camera locations are illustrated on the right side of the figure. The left side depicts the cross-camera appearance variations of two pedestrians, showcasing differences across various sessions and times of the day.}
    \label{fig:cam_views}
\end{figure*}

\section{Prior Work}
\label{sec:prior}
\textbf{Video-based Person ReID Datasets.} Existing person ReID datasets are numerous but severely lack the ability to address real-world challenges, particularly in aerial-ground scenarios. 
{Ground-based datasets like MARS~\cite{mars} and LS-VID~\cite{GLTR_LS-VID} provide large-scale benchmarks but focus mainly on ground perspectives, highlighting the need for multi-platform surveillance datasets.}
{The inclusion of clothing changes in datasets such as MEVID~\cite{Davila2023mevid}, CCVID~\cite{gu2022CAL}, and VCCR~\cite{han20223d} represents progress in addressing real-world challenges, though they could benefit from more identities and diverse environments.}
{The BRIAR dataset~\cite{cornett2023expanding}, while featuring 1,000 subjects and UAV footage, primarily targets face recognition with restricted access.} 
{P-Destre~\cite{kumar2020p} pioneered aerial view exploration, though its use of a single drone at lower altitudes (5-6m) creates opportunities for datasets covering higher operational altitudes more common in surveillance applications.}
{The G2A-VReID dataset by Zhang \emph{et al.}~\cite{zhang2024cross} represents an important step in combining aerial and ground views. While innovative, it contains 2,788 identities within a 20-60m altitude range using 2 cameras, suggesting opportunities for future datasets to expand in scale, altitude diversity, camera count, and environmental variety.}
Tab.~\ref{tab:dataset_comparison} compares our AG-VPReID dataset with others across multiple dimensions.

\vspace{3px}
\hspace{-12px}\textbf{Video-based Person ReID.} Video-based person ReID methods have evolved to leverage both spatial and temporal cues. Early approaches used recurrent neural networks and 3D convolutional networks \cite{mclaughlin2016recurrent,li2019multi}, while later works incorporated temporal pooling \cite{mars} and attention mechanisms \cite{liu2021watching}. Recent advancements include temporal complementary learning \cite{hou2020temporal}, Transformer-based architectures \cite{liu2021video,he2021dense}, and techniques addressing cross-platform and cloth-changing scenarios \cite{zhang2024cross,wang2021pyramid,han20223d}. {The AG-ReID 2023 Challenge~\cite{nguyen2023ag} highlighted aerial-ground ReID challenges, with winners employing re-ranking, data augmentation, and centroid-based representations. Recent works include Instruct-ReID~\cite{he2024instruct} with instruction-guided retrieval, Domain Shifting~\cite{jiang2025domain} for distribution adaptation, and SEAS~\cite{zhu2024seas} using 3D body shape guidance.} Despite these developments, most existing methods employ uni-modal frameworks trained on predefined label sets. In contrast, recent work has proposed a visual-language multi-modal learning paradigm \cite{yu2024tf}, potentially offering more flexibility and robustness in feature representation for video-based person ReID.

\section{AG-VPReID Dataset} 
\label{sec:dataset_collection}
This section offers a detailed overview of the creation process for our AG-VPReID dataset.
We describe the methods used for collecting video footage in Sec.~ \ref{data_collect}. Sec.~ \ref{label_process} introduces our annotation procedures. Sec.~\ref{sec:data_chars} compares AG-VPReID with existing datasets, highlighting its unique features.  

\subsection{Dataset Collection} 
\label{data_collect}
The AG-VPReID dataset was collected over a period of 20 days, including 10 morning sessions (8:30am-10:00am) and 10 afternoon sessions (3:30pm-5:00pm), with each session lasting 60 minutes. Data capture involved two drones, two CCTV cameras, and two wearable cameras.
Each drone operated at four different altitudes—15m, 30m, 80m, and 120m—for 15 minutes per session, providing a comprehensive range of aerial views. In total, the dataset comprises 240 hours of video footage, documented in Tab. \ref{tab:equipment}, which includes detailed specifications of the equipment used. 
The dataset features diverse resolutions, frame rates, and perspectives, extending from ground level to 120-meter aerial views.
Fig.~\ref{fig:cam_views} shows example frames across different viewpoints and image qualities from different platforms. The drones, CCTV, and wearable devices are positioned to view individuals from different angles, as illustrated in Fig.~\ref{fig:cam_views}, forcing person ReID models to learn robust multiview and partial-view representations to be effective.

\begin{table}[ht]
    \centering
    \scriptsize
    % \footnotesize
    \begin{tabular}{@{}l@{\hspace{2pt}}l@{\hspace{2pt}}c@{\hspace{2pt}}c@{\hspace{2pt}}c@{}}
    \toprule
    \textbf{Type} & \textbf{Model} & \textbf{Resolution} & \textbf{Lens} & \textbf{FPS} \\
    \midrule
    CCTV & Bosch (Outdoor) & $704 \times 480$ & 24mm& 15 \\
         & Bosch (Indoor)  & $1280 \times 720$ & 18mm& 25 \\
    \midrule
    Wearable & GoPro10 (Front) & $3840 \times 2160$ & 16mm & 30 \\
             & GoPro10 (Side) & $1920 \times 1080$ & 16mm & 60 \\
    \midrule
    Drones & DJI Inspire2 & $3840 \times 2160$ & 24mm & 25 \\
           & DJI M300RTK & $8192 \times 5460$ & 35mm & 1 \\
    \bottomrule
    \end{tabular}
\vspace{-6px}    \caption{\footnotesize Equipment specifications for the AG-VPReID dataset.}
    \label{tab:equipment}
\end{table}

To ensure professional drone operations, a specialized team (one RPAS engineer, one Chief Remote Pilot, one RPAS technician) managed all 20 data collection days, handling flights, capturing aerial footage, and performing initial data preprocessing.

\subsection{Labeling Process}
\label{label_process}
{The AG-VPReID dataset uses YOLOv8x for person detection and tracking \cite{Jocher2024}, extracting images from all frames across multiple cameras. It includes both short-term and long-term ReID scenarios, the latter including instances where participants change clothes to test long-term identity persistence. Identity matching was performed by a team of expert annotators, supported by research assistants, to ensure both accuracy and consistency in the matching process.
Following \cite{Nguyen2024AGReIDv2BA}, we manually annotated each identity with 15 selective soft-biometrics attributes to enhance the dataset's utility for attribute-based person ReID applications. For a detailed list of these attributes, refer to \ref{sup:softattributes}.
}

\subsection{Dataset Characteristics} \label{sec:data_chars}
Compared with existing video-based ReID datasets, our AG-VPReID dataset has five unique characteristics:

\vspace{3px}
\hspace{-12px}\textbf{1) The highest number of identities and tracklets. }  
{AG-VPReID sets a new benchmark with 6,632 unique identities and 32,321 tracklets, substantially surpassing existing datasets. 
It holds nearly twice as many identities as prominent ground-based datasets such as LSVID~\cite{GLTR_LS-VID}, which contains 3,772 identities, and more than six times the tracklets of the next largest aerial-ground dataset, G2A-VReID~\cite{zhang2024cross}, which includes 5,576 tracklets. Additionally, AG-VPReID encompasses over 9.6 million frames, providing 50 times the volume of frames compared to other aerial-ground datasets. While MEVID~\cite{Davila2023mevid} exceeds in frame count with 10.46 million, it does not match AG-VPReID in terms of identity and tracklet numbers.}

\vspace{3px}
\hspace{-12px} \textbf{2) The most diverse platforms.} 

{Our AG-VPReID dataset is the first to incorporate aerial, ground, and wearable platforms for video-based person ReID. The inclusion of wearable cameras provides a novel dimension with high-quality first-person perspectives. This combination results in extreme variations in resolution and subject size across platforms: UAV (18$\times$37 to 293$\times$542 pixels), CCTV (22$\times$23 to 172$\times$413 pixels), and GoPro (25$\times$48 to 483$\times$1078 pixels).}
 
\begin{figure}
    \centering
    \includegraphics[width=0.9\linewidth]{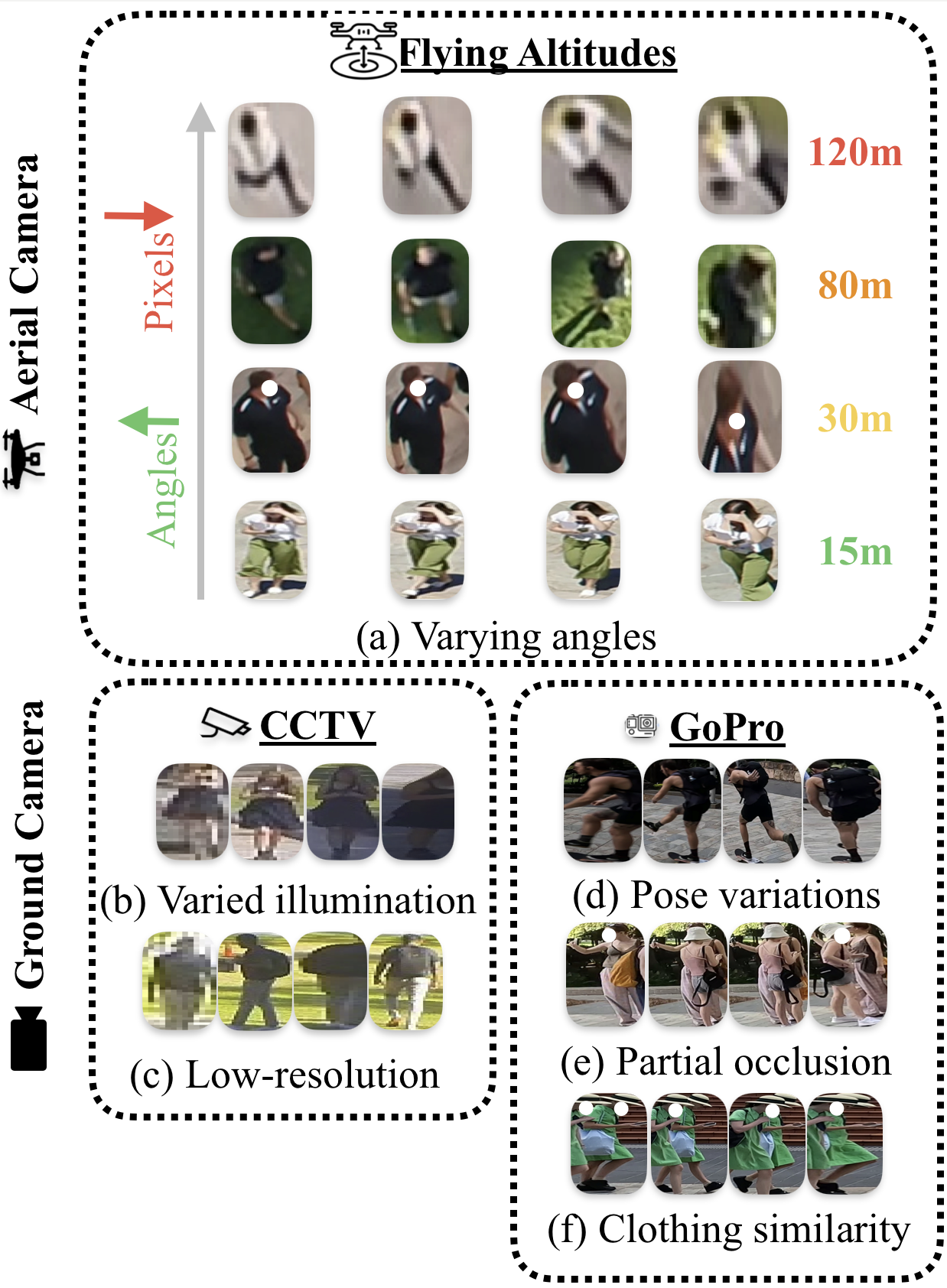}
    \caption{The AG-VPReID dataset presents several key challenges: extreme viewpoints, varying resolutions and subject sizes, pose/illumination variations, occlusions, and similar clothing among subjects.}
    \label{fig:dataset_challenges}
\end{figure}

\vspace{3px}
\hspace{-12px}\textbf{3) The highest flying altitudes.} {AG-VPReID features footage from altitudes reaching up to 120 meters, exceeding existing datasets' 60-meter maximum~\cite{zhang2024cross}. This introduces challenges: 1) extreme viewpoints with perspective distortions; 2) multiple scales with varying resolutions; and 3) image quality issues (Fig.~\ref{fig:dataset_challenges}). We used two drones—one with a wide-angle camera for area monitoring and another with a narrow-angle camera for detailed observation.}

\vspace{3px}
\hspace{-12px} \textbf{4) Rich outdoor scenarios with real-world challenges.} 
 
{Our AG-VPReID dataset presents diverse outdoor campus scenarios with real-world challenges including complex occlusions, varied poses from different activities, and uniform-wearing individuals. Fig.~\ref{fig:dataset_challenges} shows examples of these diverse scenarios.} 

\vspace{3px} 
\hspace{-12px} \textbf{5) Other notable characteristics.} {AG-VPReID includes comprehensive attributes for each identity (gender, age, clothing style, accessories), enabling fine-grained analysis. The dataset features long-term tracking data of 14 diverse individuals recorded across multiple days, each wearing different clothing per session to capture real-world variations. We also provide camera calibration information and GPS coordinates to support multi-camera tracking research. See supplementary material Sec.~\ref{sup:dataset} for details.}

% \vspace{3px} 
% \hspace{-12px}\textbf
\subsection{Ethics and Privacy}  
This research received ethics approval for data collection and usage. We implement ``Deface'' \cite{deface} to blur faces, secure data storage, and obtained informed consent from all participants. Details are available at our project \href{https://github.com/agvpreid25/AG-VPReID-Net}{repository}.

\section{AG-VPReID-Net} 
\label{sec:method}

\begin{figure*}[t]
    \centering
    \includegraphics[width=\linewidth]{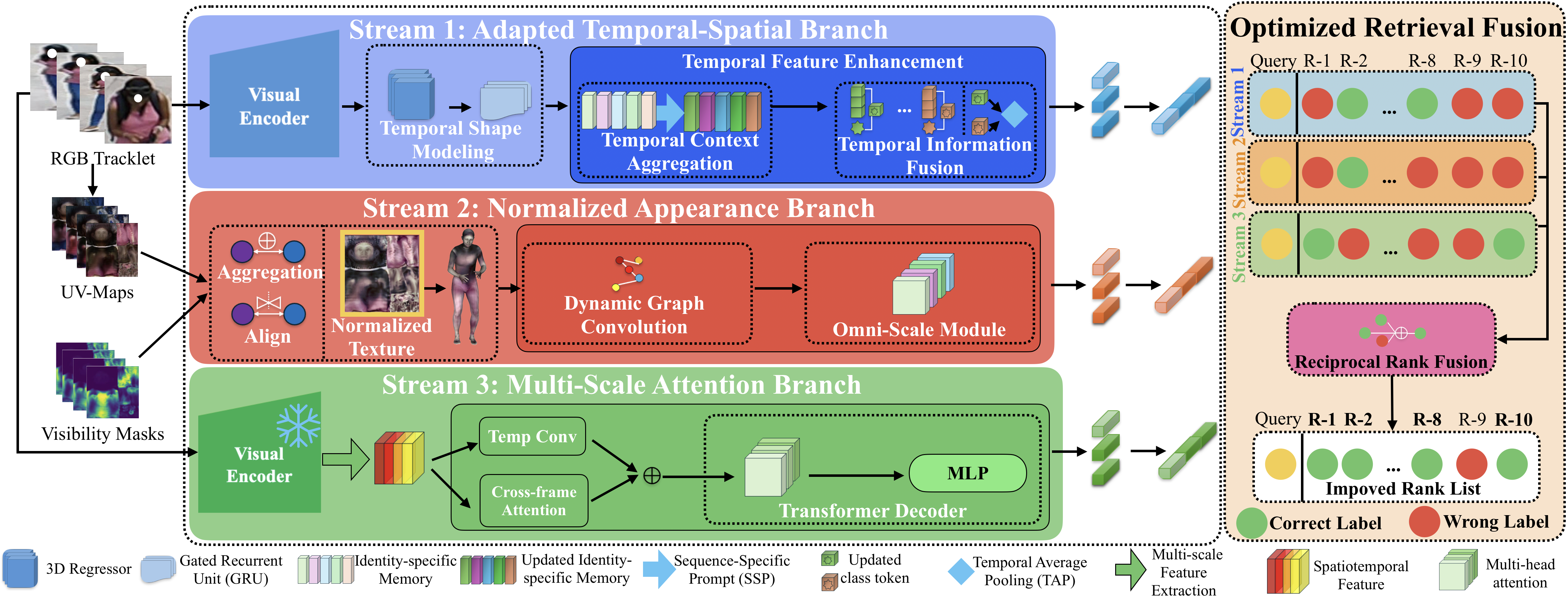}
    \caption{The three-stream AG-VPReID-Net architecture addresses aerial-ground ReID challenges: Temporal-Spatial stream for motion modeling and temporal features, Normalized Appearance for resolution/appearance variations, and Multi-Scale Attention for aerial-ground scale variations.}
    \label{fig:model-architecture}
\end{figure*}

% short
%Aerial-ground person ReID faces unique challenges including significant view changes, resolution variations, and occlusions. 
We propose AG-VPReID-Net, a purpose-built framework addressing aerial-ground ReID's unique challenges. In particular, we propose an Adapted Temporal-Spatial Stream for robust temporal-spatial representations to deal with the temporal discontinuity challenge caused by drone motion from unstable tracking between frames. We propose a Normalized Appearance Stream for resolution and appearance changes to deal with extreme viewpoint shifts. To deal with altitude-driven scale variance, we introduce a Multi-Scale Attention Stream for scale variations. Fig.~\ref{fig:model-architecture} illustrates our architecture. Detailed stream contributions are provided in Tab.~\ref{tab:model_overview} of the supplementary material.

\subsection{Stream 1: Adapted Temporal-Spatial Stream}\label{sec:ada_tem_spa_stream}

When performing video-based person ReID, a key challenge is handling inconsistent motion patterns and temporal gaps between video frames. To address this, 
we propose an Adapted Temporal-Spatial stream that combines CLIP's visual encoder with temporal and 3D shape modeling to create a comprehensive representation of individuals. 
Our method operates on a sequence $\mathcal{V}$ of $T$ frames through the following components:

\vspace{3px}
\hspace{-12px}\textbf{Visual Feature Extraction:} Using CLIP's visual encoder $E_v(\cdot)$, we extract frame-level features,
\begin{align}
F_t &= E_v(\mathcal{I}_t),  \quad t \in \{1,...,T\},
\end{align}
where $\mathcal{I}_t$ and $F_t$ are the $t$-th frame and its features.

\vspace{3px}
\hspace{-12px}\textbf{Temporal Processing:} We incorporate temporal modeling through two key components:

1) \textit{Temporal 3D Shape Modeling (TSM):} Following ~\cite{nguyen2024temporal}, we extract 3D shape representations,
\begin{equation}
g_t = \text{GRU}(F_t, g_{t-1}), 
\beta_t = \text{3D Regressor}(g_t),
\end{equation}
where $g_t$ captures temporal dynamics and $\beta_t$ represents SMPL model parameters.

2) \textit{Temporal Feature Enhancement (TFE):} Adapting from \cite{yu2024tf}, we enhance features by combining appearance and shape,
\begin{align}
F_{enhanced} &= \text{TFE}(F_{1:T}, \beta_{1:T}).
\end{align}

%\vspace{6px}
\hspace{-12px}\textbf{Identity-Aware Processing:} We incorporate identity information through,
\begin{align}
\mathcal{M}_{y_i} &= \frac{1}{N_{y_i}} \sum_{V \in \mathcal{V}_{y_i}} \text{TAP}(F_{enhanced}), \notag \\
\mathcal{M}_{y_i}^{refined} &= \text{SSP}(F_{enhanced}, \mathcal{M}_{y_i}), \notag \\
R_{stream1} &= [F_{enhanced}; \mathcal{M}_{y_i}^{refined}],
\end{align}
where $\mathcal{M}_{y_i}$ is the identity memory bank constructed through Temporal Average Pooling (TAP), and the Sequence-Specific Prompt (SSP) module refines this representation for the final output $R_{stream1}$.

\subsection{Stream 2: Normalized Appearance Stream}\label{sec:norm_app_stream}

The Adapted Temporal-Spatial (ATS) stream provides robust temporal-spatial representation, but may not fully capture fine-grained appearance details across viewpoints, especially in aerial footage. To address this limitation, we propose a Normalized Appearance (NA) stream that effectively aggregates appearance details from multiple viewpoints.

The NA stream normalizes and combines appearance information across frames using UVTexture maps and visibility masks. Our process involves: (1) Extracting UVTexture maps and visibility masks per frame, (2) Normalizing UVTexture maps brightness, (3) Aligning maps across frames, (4) Weighted aggregation using visibility masks, and (5) Generating the final normalized representation. The brightness normalization and weighted aggregation of UVTexture maps can be formulated as,
\begin{align}
    T_i^{norm} &= \gamma(H(N(T_i))), \\
    T_{aggregated} &= \frac{\sum_{i=1}^{N} V_i \odot T_i^{norm}}{\sum_{i=1}^{N} V_i},
\end{align}
where $T_i^{norm}$ is the normalized UVTexture map for frame $i$, $N(\cdot)$, $H(\cdot)$, and $\gamma(\cdot)$ are normalization, histogram matching and gamma correction functions respectively. $T_{aggregated}$ is the final aggregated map. We leverage PhysPT \cite{zhang2024physpt} for pose estimation and Texformer \cite{xu20213dTexformer} to generate UV maps from PhysPT's output 3D meshes. The maps are improved through inter-frame consistency before feeding into the DGC Omni-scale Module \cite{Zheng2020ParameterEfficientPR}.

\subsection{Stream 3: Multi-Scale Attention Stream}

While the ATS stream provides a robust temporal-spatial representation and the NA stream addresses viewpoint and occlusion challenges, aerial-ground person ReID still faces significant hurdles due to extreme scale variations between drone and ground-level footage. The first two streams effectively capture temporal dynamics, 3D shape information, and viewpoint-invariant appearance details, but they may not fully address the drastic scale differences inherent in aerial-ground scenarios. To complement the ATS stream and NA stream and address this limitation, we propose a Multi-Scale Attention (MSA) stream.

%------------------- Feng --------------------------%
%%
In detail, this stream leverages the power of frozen large vision models combined with lightweight, adaptive processing. Specifically, this stream utilizes a frozen large vision model to extract multi-scale features for video-based person ReID. By combining a lightweight Transformer decoder with a local temporal module, this approach dynamically integrates spatial and temporal information, thereby enhancing our framework's ability to accurately capture essential person-specific details.

Specifically, for each frame $\mathcal{I}_t$ within the sequence $\mathcal{V}{y_i}$, the CLIP vision encoder~\cite{radford2021learning} is employed to extract features independently. The process collects tokens from various layers at regular intervals to compile a detailed feature map that captures spatial correspondences. These frame feature maps are subsequently concatenated and assembled into a spatiotemporal feature volume $\mathbf{G}$.
Following the methods~\cite{lin2022frozen,ye2024biggait}, we integrate temporal information into this volume before processing it through a Transformer decoder. This decoder globally aggregates features across multiple layers, employing a video-level classification token as a query, with feature volumes from different layers of the backbone serving as keys and values. A linear layer then maps the output of the decoder’s final block to produce class predictions. The operational dynamics of the Transformer decoder are outlined as follows,
\begin{align}
Y_i &= \operatorname{Temp}_i([\mathbf{G}_{N-M+i, 1}, \mathbf{G}_{N-M+i, 2}, ..., \mathbf{G}_{N-M+i, T}]), \notag \\
\tilde{q}_i &= q_{i-1} + \operatorname{MHA}_i(q_{i-1}, Y_i, Y_i), \notag \\
q_i &= \tilde{q}_i + \operatorname{MLP}_i(\tilde{q}_i), \notag \\
f_{G}&= \operatorname{FC}(q_M),
\label{eqn:pose}
\end{align}
where $\mathbf{G}_{n, t}$ represents the features of frame $t$ extracted from the $n$-th layer of CLIP vision encoder. The feature volume $Y_i$, which undergoes temporal modulation, is input into the $i$-th layer of the Transformer decoder. The query token $q_i$ is incrementally refined, beginning with $q_0$ as learnable initial parameters. The final output $f_{G}$, corresponds to the final feature. The spatiotemporal decoder comprises $M$ blocks. $N$ denotes the number of encoder layers. Multi-head attention (MHA) involves query, key, and value, each of which plays a distinct role. The operator $\operatorname{Temp}(\cdot)$ is utilized to model temporal dynamics, which produces feature tokens influenced by detailed temporal information.

\section{Experimental Results} 
\label{sec:exp}

\subsection{Datasets and Evaluation Metrics}
We conducted evaluations of our method using the AG-VPReID and four established video-based person ReID datasets: iLIDS~\cite{iLIDS}, Mars~\cite{mars}, LS-VID~\cite{GLTR_LS-VID} and G2A-VReID~\cite{zhang2024cross}. 
For AG-VPReID, we used a balanced split of 3,013 identities with both ground and aerial views, dividing them equally for training and testing purposes \cite{AGReIDv1, Nguyen2024AGReIDv2BA}. Details on the training and testing configurations are provided in Table~\ref{tab:eval_dataset_summary}.
We evaluate performance using the Cumulative Matching Characteristic (CMC) at Rank-1 and the mean Average Precision (mAP).

\begin{table}[!htbp]
\centering
% \footnotesize
\scriptsize
\begin{tabular}{l|c|c|c|c}
\toprule
\textbf{Case} & \textbf{Subset} & \textbf{\# IDs} & \textbf{\# Tracklets} & \textbf{\# Images (M)} \\
\midrule
\midrule
Training& All& 1,555& 13,300& 3.85\\
\midrule
 & All & 1,456 & 13,566 &3.94 \\

& 15m & 506 & 4,907 & 1.50 \\
Testing (A2G)& 30m & 377 & 2,885 & 0.89 \\
& 80m & 356 & 2,592 & 0.69 \\
& 120m & 308 & 3,182 & 0.86 \\
\midrule
& All & 5,075* & 19,021 & 5.79 \\
& 15m & 1,403 & 6,362 & 2.14 \\
Testing (G2A)& 30m & 1,406 & 4,468 & 1.41 \\
& 80m & 1,162 & 3,866 & 1.13 \\
& 120m & 1,195 & 4,325 & 1.11 \\
\bottomrule
\end{tabular}
\vspace{-6px}
\caption{\footnotesize  Statistics of AG-VPReID dataset. A2G: aerial-to-ground, G2A: ground-to-aerial. *3,619 additional IDs as distractors.}
\label{tab:eval_dataset_summary}
\end{table}

\begin{table*}[!htbp]
\centering
\scriptsize
\begin{tabular}{l||c|c|c|c|c|c||c|c||c|c|c|c}
\toprule
&     \multicolumn{2}{c|}{MARS}& \multicolumn{2}{c|}{LS-VID}& \multicolumn{2}{c||}{iLIDS-VID} & \multicolumn{2}{c||}{G2A-VReID}&\multicolumn{4}{c}{AG-VPReID}\\
\cline{2-13}
Method &     \multicolumn{6}{c||}{Ground $\rightarrow$ Ground}& \multicolumn{2}{c||}{Ground $\rightarrow$ Aerial }&\multicolumn{2}{c|}{Aerial $\rightarrow$ Ground}& \multicolumn{2}{c}{Ground $\rightarrow$ Aerial}\\
\cline{2-13}
&    mAP& Rank-1& mAP&Rank-1& Rank-1&Rank-5& mAP&Rank-1 &mAP& Rank-1& mAP&Rank-1\\
\midrule
STMP\cite{liu2019spatial} & 72.7 & 84.4 & 39.1 & 56.8 & 84.3 & 96.8 & - & - & 50.7 & 60.3 & 45.2 & 55.8 \\
M3D\cite{li2019multi} & 74.1 & 84.4 & 40.1 & 57.7 & 74.0 & 94.3 & - & - & 52.4 & 62.6 & 47.9 & 57.3 \\
GLTR\cite{li2019global} & 78.5 & 87.0 & 44.3 & 63.1 & 86.0 & 98.0 & - & - & 55.6 & 65.8 & 50.1 & 60.5 \\
TCLNet\cite{hou2020temporal} & 85.1 & 89.8 & 70.3 & 81.5 & 86.6 & - & 65.4 & 54.7 & 57.2 & 67.9 & 52.7 & 62.4 \\
MGH\cite{yan2020learning} & 85.8 & 90.0 & 61.8 & 79.6 & 85.6 & 97.1 & 76.7 & 69.9 & 60.3 & 70.8 & 55.5 & 65.2 \\
GRL\cite{liu2021watching} & 84.8 & 91.0 & - & - & 90.4 & 98.3 & - & - & 58.7 & 68.4 & 53.9 & 63.6 \\
BiCnet-TKS\cite{hou2021bicnet} & 86.0 & 90.2 & 75.1 & 84.6 & - & - & 63.4 & 51.7 & 59.8 & 69.2 & 54.3 & 64.7 \\
CTL\cite{liu2021spatial} & 86.7 & 91.4 & - & - & 89.7 & 97.0 & - & - & 56.4 & 66.9 & 51.8 & 61.3 \\
STMN\cite{eom2021video} & 84.5 & 90.5 & 69.2 & 82.1 & - & - & 66.7 & 56.1 & 61.6 & 71.5 & 56.9 & 66.2 \\
PSTA\cite{wang2021pyramid} & 85.8 & 91.5 & - & - & 91.5 & 98.1 & - & - & 60.5 & 70.2 & 55.8 & 65.7 \\
DIL\cite{he2021dense} & 87.0 & 90.8 & - & - & 92.0 & 98.0 & - & - & 61.2 & 70.9 & 56.3 & 66.1 \\
STT\cite{zhang2021spatiotemporal} & 86.3 & 88.7 & 78.0 & 87.5 & 87.5 & 95.0 & - & - & 61.0 & 70.7 & 56.1 & 65.9 \\
TMT\cite{liu2021video} & 85.8 & 91.2 & - & - & 91.3 & 98.6 & - & - & 60.8 & 70.5 & 55.9 & 65.8 \\
CAVIT\cite{wu2022cavit} & 87.2 & 90.8 & 79.2 & 89.2 & 93.3 & 98.0 & - & - & 61.4 & 71.1 & 56.5 & 66.3 \\
SINet\cite{bai2022salient} & 86.2 & 91.0 & 79.6 & 87.4 & 92.5 & - & - & - & 61.3 & 71.0 & 56.4 & 66.2 \\
MFA\cite{gu2022motion} & 85.0 & 90.4 & 78.9 & 88.2 & 93.3 & 98.7 & - & - & 61.1 & 70.8 & 56.2 & 66.0 \\
DCCT\cite{liu2023deeply} & 87.5 & 92.3 & - & - & 91.7 & 98.6 & - & - & 61.5 & 71.2 & 56.6 & 66.4 \\
LSTRL\cite{liu2023video} & 86.8 & 91.6 & 82.4 & 89.8 & 92.2 & 98.6 & - & - & 61.7 & 71.3 & 56.7 & 66.5 \\
CLIP-ReID\cite{li2023clip} & 88.1 & 91.7 & 80.6 & 88.8 &  - & - & - & -  & 62.3 & 71.6 & 57.2 & 66.8  \\
\midrule 
\textbf{AG-VPReID-Net} & \textbf{91.5} & \textbf{93.2}& \textbf{87.3}& \textbf{93.2}& \textbf{96.3} & \textbf{99.5} & \textbf{81.3}& \textbf{73.1}& \textbf{64.0} & \textbf{71.9} & \textbf{58.0} & \textbf{75.6} \\
\bottomrule
\end{tabular}
\vspace{-6px}
\caption{  Performance comparison across datasets. \textbf{Bold} shows best results.}
\label{tab:compare_with_SOTA}
\end{table*}

\subsection{Implementation Details} 
Our pipeline leverages UV maps generated by Texformer \cite{xu20213dTexformer} using 3D human meshes from PhysPT \cite{zhang2024physpt} with refined pose estimation. The UV maps are processed through normalization, histogram matching, and gamma correction before weighted blending with visibility masks. The architecture consists of three streams: an Adapted Temporal-Spatial Stream (CLIP ViT-B/16), a Normalized Appearance Stream for 3D coordinates and UV textures, and a Multi-Scale Attention Stream (CLIP ViT-L/14). More implementation details can be found in the supplementary.

\subsection{Comparison with State-of-the-Art Methods}
We evaluate our proposed method AG-VPReID-Net against several state-of-the-art approaches across multiple video-based person ReID datasets. Tab.~\ref{tab:compare_with_SOTA} summarizes the results.

\vspace{3px}
\hspace{-12px}\textbf{Ground-to-Ground Datasets.} 
Our method achieves superior performance on MARS (91.5\% mAP, 93.2\% Rank-1), outperforming CLIP-ReID by 3.4\% mAP. On LS-VID (87.3\% mAP, 93.2\% Rank-1), we surpass LSTRL by 4.9\% mAP. For iLIDS-VID, we reach 96.3\% Rank-1, which is 3.0\% higher than MFA.

\vspace{3px}
\hspace{-12px}\textbf{Cross-Platform Datasets.} On G2A-VReID, we achieve 81.3\% mAP and 73.1\% Rank-1, surpassing MGH by 4.6\% mAP. Note that the G2A-VReID dataset only provides a ground-to-aerial testing set. For AG-VPReID, we demonstrate strong results in both Ground-to-Aerial (58.0\% mAP, 75.6\% Rank-1, exceeding CLIP-ReID by 8.8\% Rank-1) and aerial-to-ground scenarios (64.0\% mAP, 71.9\% Rank-1, surpassing CLIP-ReID by 1.7\% mAP and 0.3\% Rank-1).

\begin{table}[!htbp]
    \centering
    \fontsize{5.9}{8}\selectfont
    \begin{tabular}{l ||c|c|c||c|c|c}
    \toprule
    & \multicolumn{3}{c||}{Aerial $\rightarrow$ Ground}& \multicolumn{3}{c}{Ground $\rightarrow$ Aerial}\\
    \cline{2-7}
    Method & Rank-1& Rank-5& Rank-10& Rank-1& Rank-5& Rank-10\\
    \midrule
    \midrule
    St-1& 71.52& 80.42& 83.88& 74.80& 84.27& 86.90\\
    St-2& 58.40& 70.20& 75.80& 61.50& 73.60& 78.20\\
    St-3& 61.65& 74.53& 79.15& 67.38& 78.82& 82.3\\
    \hline
    St-12& 69.50& 78.80& 82.50& 72.80& 82.60& 85.40\\
    St-13& 71.80& 80.60& 83.91& 75.40& 84.48& 86.91\\
    St-23& 65.70& 76.55& 80.90& 70.10& 80.45& 83.91\\
    \textbf{St-123}& \textbf{71.91}& \textbf{80.67}& \textbf{83.92}& \textbf{75.57}& \textbf{84.50}& \textbf{86.92}\\
    \bottomrule
    \end{tabular}
    \vspace{-6px}
    \caption{\footnotesize Ranking accuracy (\%) improvement on AG-VPReID dataset. }
    \label{tab:method_ablation_study_ranking}
\end{table}

\begin{table}[!htbp]
    \centering
    \fontsize{5.9}{8}\selectfont
    \begin{tabular}{l ||c|c|c|c|| c|c|c|c}
    \toprule
    & \multicolumn{4}{c||}{Aerial $\rightarrow$ Ground} & \multicolumn{4}{c}{Ground $\rightarrow$ Aerial}\\
    \cline{2-9}
    Method & 15m & 30m & 80m & 120m & 15m & 30m & 80m & 120m \\
    \midrule
    \midrule
    St-1& 80.28& 78.76& 67.24& 52.47& 83.25& 83.03& 67.07&62.31\\
    St-2& 69.75& 68.13& 52.62& 38.12& 72.87& 71.41& 52.22&45.43\\
    St-3& 74.25& 74.01& 52.53& 35.13& 77.18& 78.91& 56.59&52.56\\
    \hline
    St-12& 78.32& 76.82& 65.24& 50.53& 81.34& 81.27& 65.17&60.43\\
    St-13& 80.55& 78.95& 67.50& 52.85& 83.70& 83.55& 67.60&63.10\\
    St-23& 76.45& 74.90& 57.40& 42.70& 79.55& 79.40& 57.40&52.80\\
    \textbf{St-123}& \textbf{80.66}& \textbf{79.00}& \textbf{67.63}& \textbf{53.00}& \textbf{83.92}& \textbf{83.66}& \textbf{67.82}& \textbf{63.32}\\
    \bottomrule
    \end{tabular}
    \vspace{-6px}
    \caption{Rank-1 accuracy (\%) on AG-VPReID at various altitudes.}
    \label{tab:method_ablation_study_altitude}
\end{table}

\subsection{Ablation Study}
We conduct an ablation study on AG-VPReID to evaluate each stream. St-1 is our temporal modeling stream, St-2 is the appearance normalization stream, and St-3 is the multi-scale feature stream. Their combinations (St-12/13/23/123) merge multiple streams.

\vspace{3px}
\hspace{-12px}\textbf{Stream Contributions.} 
Tab.~\ref{tab:method_ablation_study_ranking} shows that St-1 achieves the strongest individual performance (71.52\% A2G, 74.80\% G2A Rank-1). St-2 and St-3 show moderate results (58.40\% and 61.65\% A2G Rank-1). Combined streams demonstrate complementary strengths, with St-123 achieving the best results (71.91\% A2G, 75.57\% G2A Rank-1) by integrating the three streams. 
\begin{figure}
    \centering
    \includegraphics[width=0.85\linewidth]{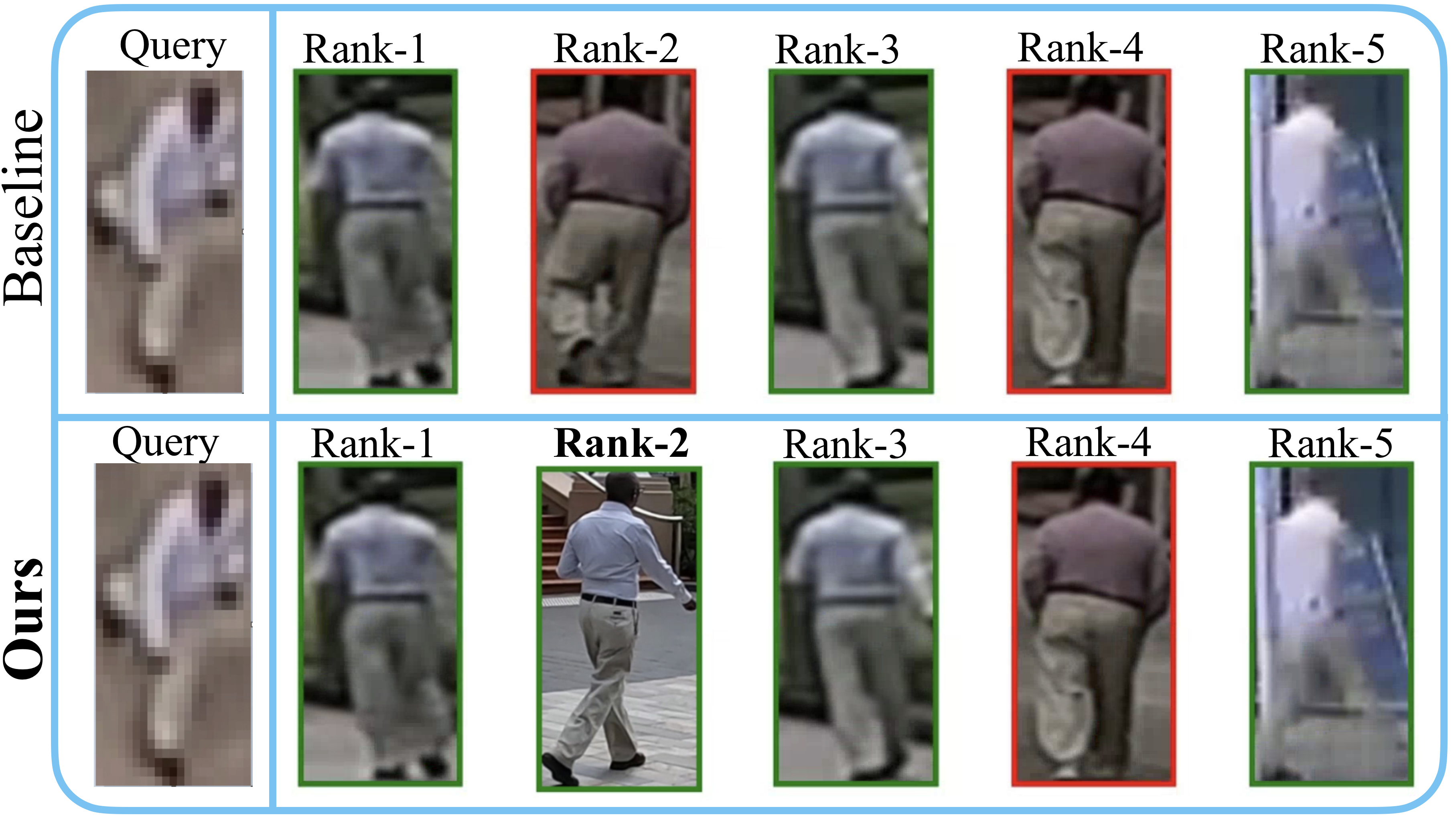}
    \caption{ Baseline vs our method on AG-VPReID dataset. Green/red: correct/incorrect labels. First tracklet image shown. Ranks show improvements in \textbf{bold}. }
    \label{fig:aerial_to_ground_120}
\end{figure}

\vspace{3px} 
\hspace{-12px}\textbf{Impact of Altitude.} 
{Table~\ref{tab:method_ablation_study_altitude} shows performance decreasing with altitude, most significantly between 30m-80m. A2G Rank-1 drops $\sim$11\% across streams. At 120m, St-1 demonstrates robustness (52.47\% vs 38.12\%/35.13\%), achieving +17.34\% improvement through temporal modeling. St-2's physics-informed UV mapping provides +7.57\% improvement (42.70\% vs. 35.13\%), while St-3's multi-scale attention yields +17.72\% improvement (52.85\% vs. 35.13\%). St-123 maintains best performance across all altitudes by combining stream strengths.}

\vspace{3px} 
\hspace{-12px}\textbf{Clothing Changes vs. Camera Angles.} 
{Analysis shows altitude increases (15m → 120m) reduce Rank-1 by 27.66\%, significantly more than clothing changes (7.85\% in ground-to-ground). Without clothing changes, aerial-ground matching (71.91\% Rank-1) still underperforms ground-to-ground (91.52\%) due to viewpoint differences. When combining aerial views with clothing changes (65.83\% Rank-1), these factors create synergistic challenges where viewpoint differences amplify clothing ambiguity. See Table~\ref{tab:impact_analysis}.}

\subsection{Visualization}\label{sec:visualization}
We further visualize the ReID results with Top-5 ranking to understand how our model improves compared to the baseline~\cite{li2023clip} for aerial-to-ground person ReID in Fig.~\ref{fig:aerial_to_ground_120}. Unlike the baseline which may be biased by image resolution and clothing textures, our approach pays attention to more robust features like motion patterns and body shape characteristics, which explains its successful identification of similar walking postures and body proportions despite the significant viewpoint differences between the aerial query and ground-view gallery pair. Additional examples are in Figs.~\ref{fig:aerial_to_ground} and~\ref{fig:ground_to_aerial} of the supplementary material.

\begin{table}[!htbp]
\centering
% \scriptsize
\fontsize{6.5}{8}\selectfont
\begin{tabular}{l|c|c|c }
\hline
\textbf{Scenario} & \textbf{Rank-1 } & \textbf{mAP } & \textbf{Key Observation}\\
\hline
\hline
\multicolumn{3}{l}{\textit{Camera Angle Impact}} \\
\hline
15m altitude (AG) & 80.66 & 77.23 & Baseline performance\\
30m altitude (AG) & 79.00 & 75.81 & 	Minimal degradation \\
80m altitude (AG) & 67.63 & 63.42 & $-13.03\%$ Rank-1 vs. 15m \\
120m altitude (AG) & \textbf{53.00} & \textbf{48.75} & $-27.66\%$ Rank-1 vs. 15m\\
\hline
\multicolumn{3}{l}{\textit{Clothing Change Impact}} \\
\hline
GG-SameClothes & 91.52 & 88.74  & Upper-bound performance\\
GG-DiffClothes & 83.67 & 79.92 & 	$-7.85\%$ Rank-1 (CC-only impact) \\
AG-SameClothes & 71.91 & 64.00 & $-19.61\%$ Rank-1 (AG-only impact)\\
AG-DiffClothes & \textbf{65.83} & \textbf{57.52} & $-6.08\%$ Rank-1 (CC impact in AG) \\
\hline
\end{tabular}
\vspace{-6px}
\caption{{Impact of clothing changes (CC) vs. camera angles.} }
\label{tab:impact_analysis}
\end{table}

\section{Conclusion} 
\label{sec:con}
We introduce AG-VPReID, a comprehensive dataset for video-based aerial-ground person ReID, addressing the critical need for a large and challenging aerial-ground dataset. We also propose AG-VPReID-Net, a purpose-built three-stream person ReID framework that combines temporal-spatial processing, physics-informed normalized appearance representation, and multi-scale attention mechanisms. This approach achieves state-of-the-art performance on both the AG-VPReID dataset and existing video-based ReID benchmarks. Notably, the relatively lower performance across all approaches on AG-VPReID highlights its demanding nature and establishes it as a robust benchmark for advancing future research in the field.

% For FINAL COPY only
\section{Acknowledgement } 
\label{sec:ack}
This work was supported by the Australian Research Council (ARC) Discovery Project (DP200101942) and a QUT Postgraduate Research Award. We gratefully acknowledge the Research Engineering Facility (REF) team at QUT for providing expertise and the research infrastructure essential for data collection and processing within this project.

{
    \small
    \bibliographystyle{ieeenat_fullname}
    \bibliography{main}

\begin{thebibliography}{53}
\providecommand{\natexlab}[1]{#1}
\providecommand{\url}[1]{\texttt{#1}}
\expandafter\ifx\csname urlstyle\endcsname\relax
  \providecommand{\doi}[1]{doi: #1}\else
  \providecommand{\doi}{doi: \begingroup \urlstyle{rm}\Url}\fi

\bibitem[Bai et~al.(2022)Bai, Ma, Chang, Huang, and Chen]{bai2022salient}
Shutao Bai, Bingpeng Ma, Hong Chang, Rui Huang, and Xilin Chen.
\newblock Salient-to-broad transition for video person re-identification.
\newblock In \emph{IEEE Conference on Computer Vision and Pattern Recognition (CVPR)}, pages 7339--7348, 2022.

\bibitem[Cornett et~al.(2023)Cornett, Brogan, Barber, Aykac, Baird, Burchfield, Dukes, Duncan, Ferrell, Goddard, et~al.]{cornett2023expanding}
David Cornett, Joel Brogan, Nell Barber, Deniz Aykac, Seth Baird, Nicholas Burchfield, Carl Dukes, Andrew Duncan, Regina Ferrell, Jim Goddard, et~al.
\newblock Expanding accurate person recognition to new altitudes and ranges: The briar dataset.
\newblock In \emph{Proceedings of the IEEE/CVF Winter Conference on Applications of Computer Vision}, pages 593--602, 2023.

\bibitem[Davila et~al.(2023)Davila, Du, Lewis, Funk, Pelt, Collins, Corona, Brown, McCloskey, Hoogs, and Clipp]{Davila2023mevid}
Daniel Davila, Dawei Du, Bryon Lewis, Christopher Funk, Joseph~Van Pelt, Roderic Collins, Kellie Corona, Matt Brown, Scott McCloskey, Anthony Hoogs, and Brian Clipp.
\newblock Mevid: Multi-view extended videos with identities for video person re-identification.
\newblock In \emph{IEEE/CVF Winter Conference on Applications of Computer Vision (WACV)}, 2023.

\bibitem[Deng et~al.(2009)Deng, Dong, Socher, Li, Li, and Fei-Fei]{deng2009imagenet}
Jia Deng, Wei Dong, Richard Socher, Li-Jia Li, Kai Li, and Li Fei-Fei.
\newblock Imagenet: A large-scale hierarchical image database.
\newblock In \emph{IEEE conference on computer vision and pattern recognition (CVPR)}, pages 248--255. Ieee, 2009.

\bibitem[Dreuw and ORB-HD(2023)]{deface}
Michael Dreuw and ORB-HD.
\newblock deface: Video anonymization by face detection, 2023.
\newblock Python package version 1.5.0.

\bibitem[Eom et~al.(2021)Eom, Lee, Lee, and Ham]{eom2021video}
Chanho Eom, Geon Lee, Junghyup Lee, and Bumsub Ham.
\newblock Video-based person re-identification with spatial and temporal memory networks.
\newblock In \emph{IEEE International Conference on Computer Vision (ICCV)}, pages 12036--12045, 2021.

\bibitem[Fu et~al.(2019)Fu, Wang, Wei, and Huang]{fu2019sta}
Yang Fu, Xiaoyang Wang, Yunchao Wei, and Thomas Huang.
\newblock Sta: Spatial-temporal attention for large-scale video-based person re-identification.
\newblock In \emph{AAAI Conference on Artificial Intelligence}, pages 8287--8294, 2019.

\bibitem[Gu et~al.(2022{\natexlab{a}})Gu, Chang, Ma, Bai, Shan, and Chen]{gu2022CAL}
Xinqian Gu, Hong Chang, Bingpeng Ma, Shutao Bai, Shiguang Shan, and Xilin Chen.
\newblock Clothes-changing person re-identification with rgb modality only.
\newblock In \emph{IEEE Conference on Computer Vision and Pattern Recognition (CVPR)}, 2022{\natexlab{a}}.

\bibitem[Gu et~al.(2022{\natexlab{b}})Gu, Chang, Ma, and Shan]{gu2022motion}
Xinqian Gu, Hong Chang, Bingpeng Ma, and Shiguang Shan.
\newblock Motion feature aggregation for video-based person re-identification.
\newblock \emph{IEEE Transactions on Image Processing}, 31:\penalty0 3908--3919, 2022{\natexlab{b}}.

\bibitem[Han et~al.(2022)Han, Huang, Gong, Wang, and Tan]{han20223d}
Ke Han, Yan Huang, Shaogang Gong, Liang Wang, and Tieniu Tan.
\newblock 3d shape temporal aggregation for video-based clothing-change person re-identification.
\newblock In \emph{Asian Conference on Computer Vision (ACCV)}, pages 2371--2387, 2022.

\bibitem[He et~al.(2021)He, Jin, Shen, Huang, Chen, and Hua]{he2021dense}
Tianyu He, Xin Jin, Xu Shen, Jianqiang Huang, Zhibo Chen, and Xian-Sheng Hua.
\newblock Dense interaction learning for video-based person re-identification.
\newblock In \emph{IEEE International Conference on Computer Vision (ICCV)}, pages 1490--1501, 2021.

\bibitem[He et~al.(2024)He, Deng, Tang, Chen, Xie, Wang, Bai, Zhu, Zhao, Ouyang, et~al.]{he2024instruct}
Weizhen He, Yiheng Deng, Shixiang Tang, Qihao Chen, Qingsong Xie, Yizhou Wang, Lei Bai, Feng Zhu, Rui Zhao, Wanli Ouyang, et~al.
\newblock Instruct-reid: A multi-purpose person re-identification task with instructions.
\newblock In \emph{Proceedings of the IEEE/CVF Conference on Computer Vision and Pattern Recognition}, pages 17521--17531, 2024.

\bibitem[Hou et~al.(2020)Hou, Chang, Ma, Shan, and Chen]{hou2020temporal}
Ruibing Hou, Hong Chang, Bingpeng Ma, Shiguang Shan, and Xilin Chen.
\newblock Temporal complementary learning for video person re-identification.
\newblock In \emph{European Conference on Computer Vision (ECCV)}, pages 388--405, 2020.

\bibitem[Hou et~al.(2021)Hou, Chang, Ma, Huang, and Shan]{hou2021bicnet}
Ruibing Hou, Hong Chang, Bingpeng Ma, Rui Huang, and Shiguang Shan.
\newblock Bicnet-tks: Learning efficient spatial-temporal representation for video person re-identification.
\newblock In \emph{IEEE Conference on Computer Vision and Pattern Recognition (CVPR)}, pages 2014--2023, 2021.

\bibitem[Jiang et~al.(2025)Jiang, Cheng, Yu, Liu, Chen, and Zhao]{jiang2025domain}
Yan Jiang, Xu Cheng, Hao Yu, Xingyu Liu, Haoyu Chen, and Guoying Zhao.
\newblock Domain shifting: A generalized solution for heterogeneous cross-modality person re-identification.
\newblock In \emph{European Conference on Computer Vision}, pages 289--306. Springer, 2025.

\bibitem[Jocher et~al.(2024)Jocher, Ayush, and Qiu]{Jocher2024}
Glenn Jocher, Ayush, and Jing Qiu.
\newblock Ultralytics {YOLO}.
\newblock \url{https://github.com/ultralytics/ultralytics}, 2024.
\newblock Accessed: 2024-03-22.

\bibitem[Kumar et~al.(2020)Kumar, Yaghoubi, Das, Harish, and Proen{\c{c}}a]{kumar2020p}
SV~Aruna Kumar, Ehsan Yaghoubi, Abhijit Das, BS Harish, and Hugo Proen{\c{c}}a.
\newblock The p-destre: A fully annotated dataset for pedestrian detection, tracking, and short/long-term re-identification from aerial devices.
\newblock \emph{IEEE Transactions on Information Forensics and Security}, 16:\penalty0 1696--1708, 2020.

\bibitem[Li et~al.(2019{\natexlab{a}})Li, Wang, Tian, Gao, and Zhang]{GLTR_LS-VID}
J. Li, J. Wang, Q. Tian, W. Gao, and S. Zhang.
\newblock Global-local temporal representations for video person re-identification.
\newblock In \emph{IEEE International Conference on Computer Vision (ICCV)}, pages 3958--3967, 2019{\natexlab{a}}.

\bibitem[Li et~al.(2019{\natexlab{b}})Li, Wang, Tian, Gao, and Zhang]{li2019global}
Jianing Li, Jingdong Wang, Qi Tian, Wen Gao, and Shiliang Zhang.
\newblock Global-local temporal representations for video person re-identification.
\newblock In \emph{IEEE International Conference on Computer Vision (ICCV)}, pages 3958--3967, 2019{\natexlab{b}}.

\bibitem[Li et~al.(2019{\natexlab{c}})Li, Zhang, and Huang]{li2019multi}
Jianing Li, Shiliang Zhang, and Tiejun Huang.
\newblock Multi-scale 3d convolution network for video based person re-identification.
\newblock In \emph{AAAI Conference on Artificial Intelligence}, pages 8618--8625, 2019{\natexlab{c}}.

\bibitem[Li et~al.(2023)Li, Sun, and Li]{li2023clip}
Siyuan Li, Li Sun, and Qingli Li.
\newblock Clip-reid: exploiting vision-language model for image re-identification without concrete text labels.
\newblock In \emph{AAAI Conference on Artificial Intelligence}, pages 1405--1413, 2023.

\bibitem[Lin et~al.(2019)Lin, Zheng, Zheng, Wu, and Yang]{Lin2019ImprovingPR}
Yutian Lin, Liang Zheng, Zhedong Zheng, Yu Wu, and Yi Yang.
\newblock Improving person re-identification by attribute and identity learning.
\newblock \emph{ArXiv}, abs/1703.07220, 2019.

\bibitem[Lin et~al.(2022)Lin, Geng, Zhang, Gao, De~Melo, Wang, Dai, Qiao, and Li]{lin2022frozen}
Ziyi Lin, Shijie Geng, Renrui Zhang, Peng Gao, Gerard De~Melo, Xiaogang Wang, Jifeng Dai, Yu Qiao, and Hongsheng Li.
\newblock Frozen clip models are efficient video learners.
\newblock In \emph{European Conference on Computer Vision (ECCV)}, pages 388--404. Springer, 2022.

\bibitem[Liu et~al.(2021{\natexlab{a}})Liu, Zha, Wu, Zheng, and Sun]{liu2021spatial}
Jiawei Liu, Zheng-Jun Zha, Wei Wu, Kecheng Zheng, and Qibin Sun.
\newblock Spatial-temporal correlation and topology learning for person re-identification in videos.
\newblock In \emph{IEEE Conference on Computer Vision and Pattern Recognition (CVPR)}, pages 4370--4379, 2021{\natexlab{a}}.

\bibitem[Liu et~al.(2021{\natexlab{b}})Liu, Zhang, Yu, Lu, and Yang]{liu2021watching}
Xuehu Liu, Pingping Zhang, Chenyang Yu, Huchuan Lu, and Xiaoyun Yang.
\newblock Watching you: Global-guided reciprocal learning for video-based person re-identification.
\newblock In \emph{IEEE Conference on Computer Vision and Pattern Recognition (CVPR)}, pages 13334--13343, 2021{\natexlab{b}}.

\bibitem[Liu et~al.(2023)Liu, Zhang, and Lu]{liu2023video}
Xuehu Liu, Pingping Zhang, and Huchuan Lu.
\newblock Video-based person re-identification with long short-term representation learning.
\newblock \emph{arXiv preprint arXiv:2308.03703}, 2023.

\bibitem[Liu et~al.(2024{\natexlab{a}})Liu, Yu, Zhang, and Lu]{liu2023deeply}
Xuehu Liu, Chenyang Yu, Pingping Zhang, and Huchuan Lu.
\newblock Deeply coupled convolution–transformer with spatial–temporal complementary learning for video-based person re-identification.
\newblock \emph{IEEE Transactions on Neural Networks and Learning Systems}, 35\penalty0 (10):\penalty0 13753--13763, 2024{\natexlab{a}}.

\bibitem[Liu et~al.(2024{\natexlab{b}})Liu, Zhang, Yu, Qian, Yang, and Lu]{liu2021video}
Xuehu Liu, Pingping Zhang, Chenyang Yu, Xuesheng Qian, Xiaoyun Yang, and Huchuan Lu.
\newblock A video is worth three views: Trigeminal transformers for video-based person re-identification.
\newblock \emph{IEEE Transactions on Intelligent Transportation Systems}, 25\penalty0 (9):\penalty0 12818--12828, 2024{\natexlab{b}}.

\bibitem[Liu et~al.(2019)Liu, Yuan, Zhou, and Li]{liu2019spatial}
Yiheng Liu, Zhenxun Yuan, Wengang Zhou, and Houqiang Li.
\newblock Spatial and temporal mutual promotion for video-based person re-identification.
\newblock In \emph{AAAI Conference on Artificial Intelligence}, pages 8786--8793, 2019.

\bibitem[McLaughlin et~al.(2016)McLaughlin, Martinez~del Rincon, and Miller]{mclaughlin2016recurrent}
Niall McLaughlin, Jesus Martinez~del Rincon, and Paul Miller.
\newblock Recurrent convolutional network for video-based person re-identification.
\newblock In \emph{IEEE Conference on Computer Vision and Pattern Recognition (CVPR)}, pages 1325--1334, 2016.

\bibitem[Nguyen et~al.(2023{\natexlab{a}})Nguyen, Nguyen, Sridharan, and Fookes]{AGReIDv1}
Huy Nguyen, Kien Nguyen, Sridha Sridharan, and Clinton Fookes.
\newblock Aerial-ground person re-id.
\newblock In \emph{IEEE International Conference on Multimedia and Expo (ICME)}, pages 2585--2590, 2023{\natexlab{a}}.

\bibitem[Nguyen et~al.(2024{\natexlab{a}})Nguyen, Nguyen, Sridharan, and Fookes]{Nguyen2024AGReIDv2BA}
Huy Nguyen, Kien Nguyen, Sridha Sridharan, and Clinton Fookes.
\newblock Ag-reid.v2: Bridging aerial and ground views for person re-identification.
\newblock \emph{IEEE Transactions on Information Forensics and Security}, 19:\penalty0 2896--2908, 2024{\natexlab{a}}.

\bibitem[Nguyen et~al.(2023{\natexlab{b}})Nguyen, Fookes, Sridharan, Liu, Liu, Ross, Michalski, Nguyen, Deb, Kothari, et~al.]{nguyen2023ag}
Kien Nguyen, Clinton Fookes, Sridha Sridharan, Feng Liu, Xiaoming Liu, Arun Ross, Dana Michalski, Huy Nguyen, Debayan Deb, Mahak Kothari, et~al.
\newblock Ag-reid 2023: Aerial-ground person re-identification challenge results.
\newblock In \emph{2023 IEEE International Joint Conference on Biometrics (IJCB)}, pages 1--10. IEEE, 2023{\natexlab{b}}.

\bibitem[Nguyen et~al.(2024{\natexlab{b}})Nguyen, Mantini, and Shah]{nguyen2024temporal}
Vuong~D Nguyen, Pranav Mantini, and Shishir~K Shah.
\newblock Temporal 3d shape modeling for video-based cloth-changing person re-identification.
\newblock In \emph{IEEE/CVF Winter Conference on Applications of Computer Vision (WACV)}, pages 173--182, 2024{\natexlab{b}}.

\bibitem[Pan et~al.(2023)Pan, Liu, Chen, He, Zheng, Zheng, and He]{pan2023pose}
Honghu Pan, Qiao Liu, Yongyong Chen, Yunqi He, Yuan Zheng, Feng Zheng, and Zhenyu He.
\newblock Pose-aided video-based person re-identification via recurrent graph convolutional network.
\newblock \emph{IEEE Transactions on Circuits and Systems for Video Technology}, 33\penalty0 (12):\penalty0 7183--7196, 2023.

\bibitem[Radford et~al.(2021)Radford, Kim, Hallacy, Ramesh, Goh, Agarwal, Sastry, Askell, Mishkin, Clark, et~al.]{radford2021learning}
Alec Radford, Jong~Wook Kim, Chris Hallacy, Aditya Ramesh, Gabriel Goh, Sandhini Agarwal, Girish Sastry, Amanda Askell, Pamela Mishkin, Jack Clark, et~al.
\newblock Learning transferable visual models from natural language supervision.
\newblock In \emph{International Conference on Machine Learning (ICML)}, pages 8748--8763, 2021.

\bibitem[Thanh et~al.(2022)Thanh, Fookes, Sridharan, Tian, Liu, Liu, and Ross]{ASAS}
Kien~Nguyen Thanh, Clinton Fookes, Sridha Sridharan, Yingli Tian, Feng Liu, Xiaoming Liu, and Arun Ross.
\newblock The state of aerial surveillance: {A} survey.
\newblock \emph{CoRR}, abs/2201.03080, 2022.

\bibitem[Wang and Zhao(2014)]{iLIDS}
Xiaogang Wang and Rui Zhao.
\newblock Person re-identification: System design and evaluation overview.
\newblock In \emph{Person Re-Identification}, pages 351--370. Springer, 2014.

\bibitem[Wang et~al.(2021)Wang, Zhang, Gao, Geng, Lu, and Wang]{wang2021pyramid}
Yingquan Wang, Pingping Zhang, Shang Gao, Xia Geng, Hu Lu, and Dong Wang.
\newblock Pyramid spatial-temporal aggregation for video-based person re-identification.
\newblock In \emph{IEEE Conference on Computer Vision and Pattern Recognition (CVPR)}, pages 12026--12035, 2021.

\bibitem[Wu et~al.(2022)Wu, He, Liu, Yang, Lei, Mei, and Li]{wu2022cavit}
Jinlin Wu, Lingxiao He, Wu Liu, Yang Yang, Zhen Lei, Tao Mei, and Stan~Z Li.
\newblock Cavit: Contextual alignment vision transformer for video object re-identification.
\newblock In \emph{European Conference on Computer Vision (ECCV)}, pages 549--566. Springer, 2022.

\bibitem[Xu and Loy(2021)]{xu20213dTexformer}
Xiangyu Xu and Chen~Change Loy.
\newblock 3d human texture estimation from a single image with transformers.
\newblock In \emph{IEEE International Conference on Computer Vision (ICCV)}, pages 13849--13858, 2021.

\bibitem[Xu et~al.(2020)Xu, Chen, Moreno-Noguer, Jeni, and De~la Torre]{xu20203d}
Xiangyu Xu, Hao Chen, Francesc Moreno-Noguer, L{\'a}szl{\'o}~A Jeni, and Fernando De~la Torre.
\newblock 3d human shape and pose from a single low-resolution image with self-supervised learning.
\newblock In \emph{European Conference on Computer Vision (ECCV)}, pages 284--300. Springer, 2020.

\bibitem[Yan et~al.(2020)Yan, Qin, Chen, Liu, Zhu, Tai, and Shao]{yan2020learning}
Yichao Yan, Jie Qin, Jiaxin Chen, Li Liu, Fan Zhu, Ying Tai, and Ling Shao.
\newblock Learning multi-granular hypergraphs for video-based person re-identification.
\newblock In \emph{IEEE Conference on Computer Vision and Pattern Recognition (CVPR)}, pages 2899--2908, 2020.

\bibitem[Ye et~al.(2024)Ye, Fan, Ma, Liu, and Yu]{ye2024biggait}
Dingqiang Ye, Chao Fan, Jingzhe Ma, Xiaoming Liu, and Shiqi Yu.
\newblock Biggait: Learning gait representation you want by large vision models.
\newblock In \emph{IEEE/CVF Conference on Computer Vision and Pattern Recognition (CVPR)}, pages 200--210, 2024.

\bibitem[Yu et~al.(2024)Yu, Liu, Wang, Zhang, and Lu]{yu2024tf}
Chenyang Yu, Xuehu Liu, Yingquan Wang, Pingping Zhang, and Huchuan Lu.
\newblock Tf-clip: Learning text-free clip for video-based person re-identification.
\newblock In \emph{AAAI Conference on Artificial Intelligence}, pages 6764--6772, 2024.

\bibitem[Zhang et~al.(2024{\natexlab{a}})Zhang, Wang, Patel, Xie, and Lai]{CARGO}
Quan Zhang, Lei Wang, Vishal~M. Patel, Xiaohua Xie, and Jianhaung Lai.
\newblock View-decoupled transformer for person re-identification under aerial-ground camera network.
\newblock In \emph{IEEE/CVF Conference on Computer Vision and Pattern Recognition (CVPR)}, pages 22000--22009, 2024{\natexlab{a}}.

\bibitem[Zhang et~al.(2024{\natexlab{b}})Zhang, Luo, Cheng, Yang, Ran, Xing, and Zhang]{zhang2024cross}
Shizhou Zhang, Wenlong Luo, De Cheng, Qingchun Yang, Lingyan Ran, Yinghui Xing, and Yanning Zhang.
\newblock Cross-platform video person reid: A new benchmark dataset and adaptation approach.
\newblock In \emph{European Conference on Computer Vision (ECCV)}, 2024{\natexlab{b}}.

\bibitem[Zhang et~al.(2021)Zhang, Wei, Xie, Zhuang, Zhang, Li, and Tian]{zhang2021spatiotemporal}
Tianyu Zhang, Longhui Wei, Lingxi Xie, Zijie Zhuang, Yongfei Zhang, Bo Li, and Qi Tian.
\newblock Spatiotemporal transformer for video-based person re-identification.
\newblock \emph{arXiv:2103.16469}, 2021.

\bibitem[Zhang et~al.(2024{\natexlab{c}})Zhang, Kephart, Cui, and Ji]{zhang2024physpt}
Yufei Zhang, Jeffrey~O Kephart, Zijun Cui, and Qiang Ji.
\newblock Physpt: Physics-aware pretrained transformer for estimating human dynamics from monocular videos.
\newblock In \emph{IEEE/CVF Conference on Computer Vision and Pattern Recognition (CVPR)}, pages 2305--2317, 2024{\natexlab{c}}.

\bibitem[Zhang et~al.(2020)Zhang, Lan, Zeng, and Chen]{zhang2020multi}
Zhizheng Zhang, Cuiling Lan, Wenjun Zeng, and Zhibo Chen.
\newblock Multi-granularity reference-aided attentive feature aggregation for video-based person re-identification.
\newblock In \emph{IEEE Conference on Computer Vision and Pattern Recognition (CVPR)}, pages 10407--10416, 2020.

\bibitem[Zheng et~al.(2016)Zheng, Bie, Sun, Wang, Su, Wang, and Tian]{mars}
L. Zheng, Z. Bie, Y. Sun, J. Wang, C. Su, S. Wang, and Q. Tian.
\newblock Mars: A video benchmark for large-scale person re-identification.
\newblock In \emph{European Conference on Computer Vision (ECCV}, pages 868--884, 2016.

\bibitem[Zheng et~al.(2022)Zheng, Wang, Zheng, and Yang]{Zheng2020ParameterEfficientPR}
Zhedong Zheng, Xiaohan Wang, Nenggan Zheng, and Yi Yang.
\newblock Parameter-efficient person re-identification in the 3d space.
\newblock \emph{IEEE Transactions on Neural Networks and Learning Systems}, 35\penalty0 (6):\penalty0 7534--7547, 2022.

\bibitem[Zhu et~al.(2024)Zhu, Budhwant, Zheng, and Nevatia]{zhu2024seas}
Haidong Zhu, Pranav Budhwant, Zhaoheng Zheng, and Ram Nevatia.
\newblock Seas: Shape-aligned supervision for person re-identification.
\newblock In \emph{Proceedings of the IEEE/CVF Conference on Computer Vision and Pattern Recognition}, pages 164--174, 2024.

\end{thebibliography}
}

\clearpage
\setcounter{page}{1}
\maketitlesupplementary

\section{Dataset}
\label{sup:dataset}
\subsection{Soft-biometric Attributes}\label{sup:softattributes}

\begin{figure}[!htbp]
    \centering
    \includegraphics[width=1\linewidth]{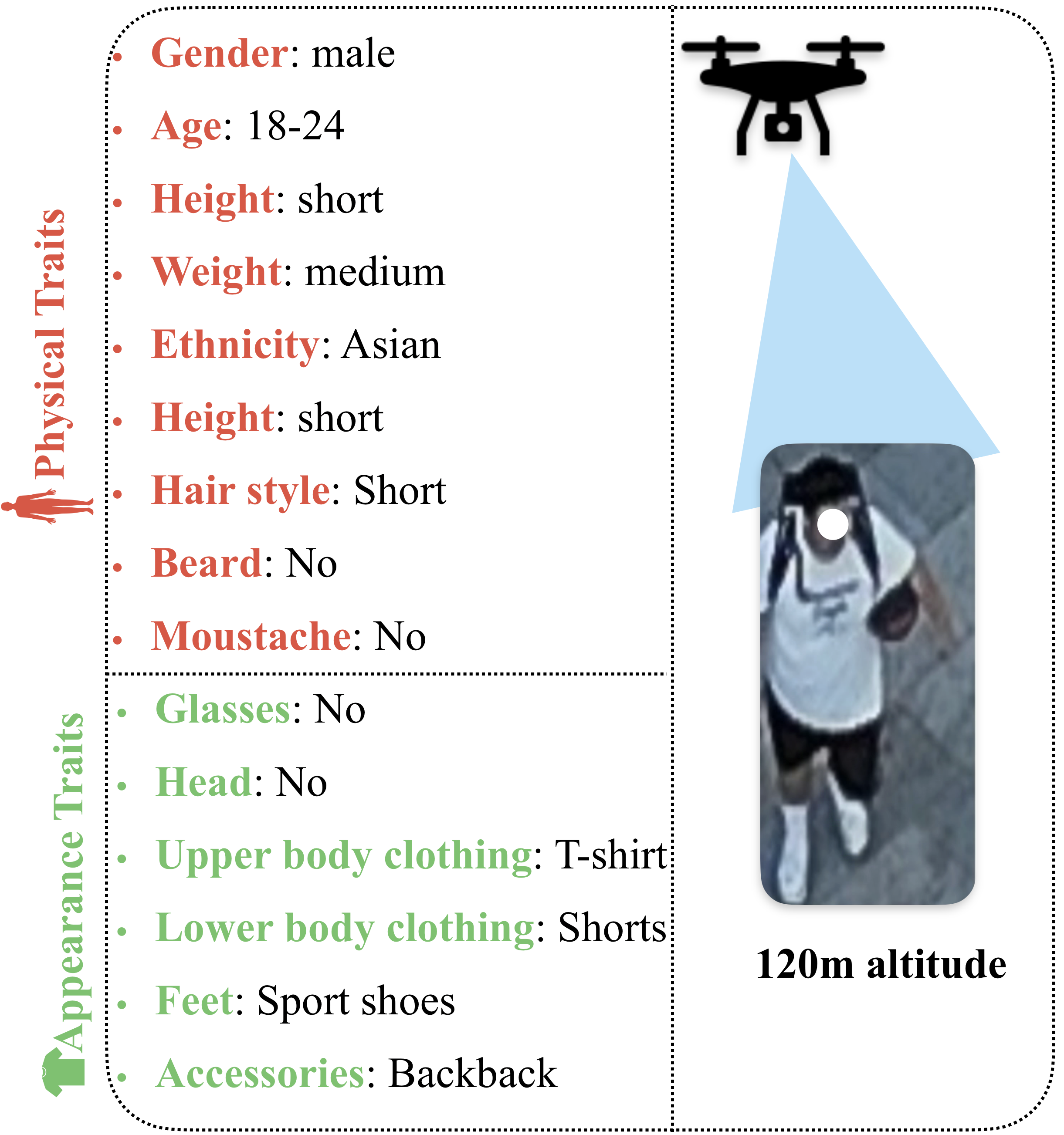}
    \caption{Soft-biometric attributes in our AG-VPReID dataset, showing Physical (top) and Appearance (bottom) Traits of a person from a top view. The attributes are categorized into physical characteristics (such as gender, age, height) and appearance details (such as clothing and accessories).}
    \label{fig:attributes}
\end{figure}

\begin{figure}[!htbp]
    \centering
    \includegraphics[width=1\linewidth]{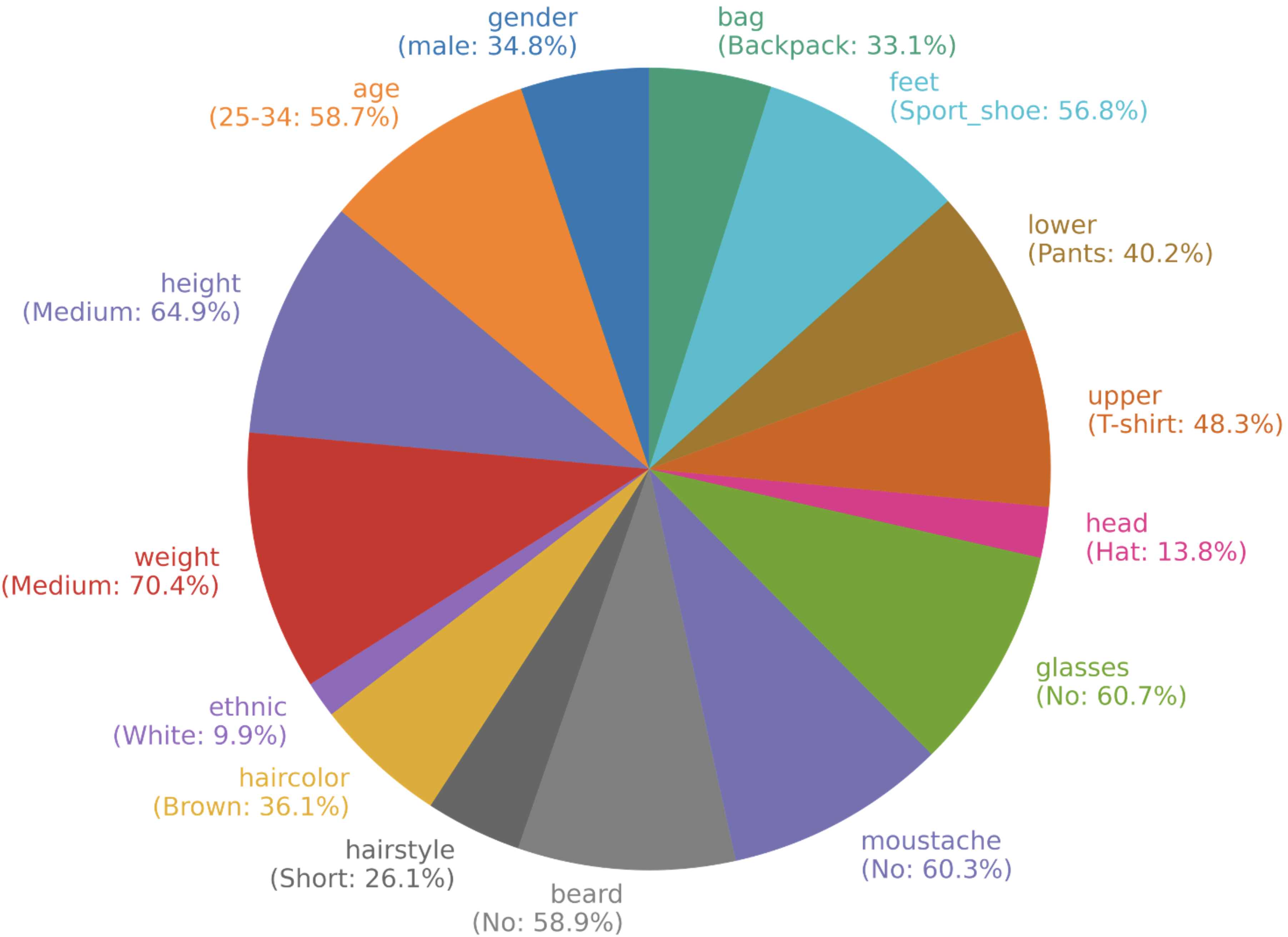}
    \caption{Most common soft-biometric attributes in our dataset. }
    \label{fig:attribute_stats}
\end{figure}

Following \cite{AGReIDv1,Nguyen2024AGReIDv2BA}, a comprehensive framework of 15 soft biometric attributes (Fig.~\ref{fig:attributes}) is employed to facilitate cross-view person identification. These attributes are categorized as physical traits (gender, age, height, weight, ethnicity, hairstyle, beard, and mustache) and appearance traits (glasses, head accessories, upper/lower body clothing, footwear, and accessories). This selection builds upon prior surveillance research \cite{kumar2020p,Lin2019ImprovingPR}, prioritizing characteristics that are both discriminative and consistent across aerial and ground viewpoints for practical real-world challenges. The distribution of these soft-biometric attributes is illustrated in Fig.~\ref{fig:attribute_stats}.

\subsection{Long-term Re-identification}
To capture realistic long-term appearance variations, we recruited 14 volunteers (consisting of nine males and five females) to participate in our data collection over a period of eight weeks with non-consecutive recording sessions. Each participant attended multiple recording sessions, with an average of four sessions per person, deliberately changing their attire between sessions. The clothing changes included variations in style (e.g., formal wear, casual wear, athletic wear), color schemes, and outer layers (e.g., jackets, coats). We instructed participants to wear clothing from their personal wardrobes to ensure naturalistic appearance variations. The recording sessions were scheduled at different times of day and under varying weather conditions, adding environmental diversity to our dataset. Each participant's sessions were separated by a minimum interval of 14 days to maximize clothing variation and capture realistic long-term appearance changes. 

\begin{table*}[!htbp]
\centering
\footnotesize
\begin{tabular}{ |c|c|c|c|c|c|c|c|c|c| }
\hline
Camera & $f_x$ & $f_y$ & $c_x$ & $c_y$ & $k_1$ & $k_2$ & $p_1$ & $p_2$ & Location \\
\hline
Bosch Outdoor & 16123.85 & 16123.85 & 298.39 & 425.13 & -0.89 & 1.06 & 0.73 & 1.79 & (27°28'36"S, 153°01'45"E) \\
Bosch Indoor & 17129.38 & 17129.38 & 306.20 & 424.81 & -1.27 & 1.27 & 0.71 & 2.06 & (27°28'40"S, 153°01'44"E) \\
GoPro10 & 17141.64 & 17141.64 & 303.62 & 401.73 & -1.66 & 2.15 & 0.48 & 1.61 & (27°28'35"S, 153°01'44"E) \\
GoPro10 & 16843.23 & 16843.23 & 264.75 & 428.89 & -1.40 & 2.03 & 0.14 & 1.26 & (27°28'35"S, 153°01'44"E) \\
DJI Inspire2 & 16467.97 & 16467.97 & 302.48 & 372.46 & -0.93 & 1.14 & 0.41 & 1.68 & (27°28'37"S, 153°01'46"E) \\
DJI M300RTK & 16837.10 & 16837.10 & 291.64 & 418.17 & -0.94 & 1.20 & 0.38 & 1.96 & (27°28'40"S, 153°01'46"E) \\
\hline
\end{tabular}
\caption{Intrinsic camera parameters and GPS coordinates}
\label{tab:intrinsic}
\end{table*}

\subsection{Calibration}

The intrinsic camera parameters, including focal length ($f_x$, $f_y$), principal point ($c_x$, $c_y$), distortion coefficients ($k_1$, $k_2$, $p_1$, $p_2$), and GPS coordinates, are presented in Tab.~\ref{tab:intrinsic}. Relative camera locations and viewing angles are visualized in Fig.~\ref{fig:cam_views} of the main paper. The camera models, resolutions, lenses, and frame rates are listed in Tab.~\ref{tab:equipment} of the main paper.

\begin{table*}[!htbp]
\centering
\begin{tabular}{|p{2.5cm}|p{6cm}|p{6cm}|}
\hline
\textbf{Stream} & \textbf{Focus} & \textbf{Challenges Addressed} \\
\hline
1. Adapted Temporal-Spatial Stream & 
\begin{itemize}
    \item Temporal shape modeling
    \item Identity-specific memory
    \item Pre-trained large vision models
\end{itemize} & 
\begin{itemize}
    \item Motion pattern inconsistencies
    \item Temporal discontinuity 
    \item Sequential feature learning
\end{itemize} \\
\hline
2. Normalized Appearance Stream & 
\begin{itemize}
    \item UV map-based appearance normalization
    \item Physics-informed techniques
    \item 3D appearance representation
\end{itemize} & 
\begin{itemize}
    \item Drastic changes in resolution
    \item Appearance variations between aerial and ground views
    \item Pose variations and partial occlusions
\end{itemize} \\
\hline
3. Multi-Scale Attention Stream & 
\begin{itemize}
    \item Multi-scale feature extraction
    \item Transformer decoder
    \item Local temporal module
\end{itemize} & 
\begin{itemize}
    \item Scale variations due to varying drone altitudes
    \item Integration of spatial and temporal information
    \item Fine-grained detail capture
\end{itemize} \\
\hline
\textbf{Overall Framework} & 
\begin{itemize}
    \item Combination of all three streams
    \item Cross-platform visual-semantic alignment
\end{itemize} & 
\begin{itemize}
    \item Robust person representation across different views
    \item Addresses viewpoint, resolution, scale, and occlusion
\end{itemize} \\
\hline
\end{tabular}
\caption{Overview of AG-VPReID-Net streams and their contributions}
\label{tab:model_overview}
\end{table*}

\section{Approach}\label{sup:approach}

\subsection{AG-VPReID-Net Framework Overview}
The AG-VPReID-Net framework, as detailed in Table~\ref{tab:model_overview}, provides a comprehensive approach to aerial-ground video person re-identification. This framework comprises three specialized streams, each designed to address specific challenges. The Adapted Temporal-Spatial Stream focuses on motion patterns and sequential feature learning, while the Normalized Appearance Stream employs UV map-based normalization and 3D appearance representation. The Multi-Scale Attention Stream utilizes multi-scale feature extraction and a transformer decoder. These streams combine to form a robust solution. This solution handles the complex challenges of matching individuals across aerial and ground-based video footage, including viewpoint variations, occlusions, and scale differences.

\subsection{Optimization}\label{sec:optimization}

Our AG-VPReID-Net achieves optimal performance by integrating three streams. Initially, each stream produces independent feature representations. These are then combined using an adaptive weighted fusion mechanism:

\begin{equation}
F_{combined} = \alpha F_{temporal} + \beta F_{appearance} + \gamma F_{multiscale},
\end{equation}

where $\alpha$, $\beta$, and $\gamma$ are learnable parameters that adapt to the input characteristics. This allows the model to dynamically emphasize different streams depending on the specific aerial-ground matching scenario.

To further improve ranking performance, we employ Reciprocal Rank Fusion (RRF) as a post-processing step. RRF combines the individual rankings from each stream to produce a final, more robust ranking:

\begin{equation}
RRF(d) = \sum_{i=1}^{3} \frac{1}{k + r_i(d)},
\end{equation}

where $d$ is a candidate match, $k$ is a constant set to 60 in our experiments, and $r_i(d)$ is the rank of $d$ in the $i$-th stream. This fusion technique gives higher weight to candidates that rank highly across multiple streams, leading to improved mean Average Precision (mAP) and Rank-k accuracy in our experiments.

RRF is chosen for its robustness, handling cases where correct matches may have inconsistent rankings across streams. This is particularly important in aerial-ground ReID, where different streams may excel under different conditions (e.g., varying altitudes or occlusions). By combining rankings rather than raw scores, RRF provides a more robust final ranking that is less sensitive to individual stream failures.

\section{Implementation Details}
\label{sup:implementation}

\textbf{UV Map Acquisition and Processing.} Our implementation pipeline uses UV maps from Texformer \cite{xu20213dTexformer}, which utilizes 3D human meshes from RSC-Net \cite{xu20203d}. To preserve temporal relationships within individual video sequences, we integrate PhysPT \cite{zhang2024physpt} for more precise pose estimations. These refined poses are fed into Texformer, yielding higher-fidelity UV maps. We enhance inter-frame consistency through normalization, histogram matching, and gamma correction. The final UV map is constructed via weighted blending, combining processed UV maps with a visibility mask M = max(dot(N, V), 0), where N is the surface normal and V is the view vector. Blending weights are determined using a softmax function over mask values, ensuring smooth transitions between UV map regions.

\vspace{6px}
\hspace{-12px}\textbf{Configuration for Stream 1: Adapted Temporal-Spatial Stream.} The Adapted Temporal-Spatial Stream utilizes a pre-trained CLIP ViT-B/16 model as the visual encoder (frozen during training), complemented by a Temporal Shape Modeling (TSM) branch with 2 GRU layers ($1024$ neurons each) and an identity-aware 3D regressor. An Attention-based Shape Aggregation (ASA) module, consisting of 2 GRU layers and a self-attention mechanism, processes shape information. Temporal features are captured using a Temporal Memory Diffusion (TMD) module with multi-head self-attention. The stream is trained on 8-frame clips ($256 \times 128$ resolution) with a batch size of $16$, using Adam optimizer, an initial learning rate of $5 \times 10^{-3}$ with warm-up and decay, and a weight decay of $0.01$. Data augmentation includes random horizontal flipping and random erasing.

\vspace{6px}
\hspace{-12px}\textbf{Configuration for Stream 2: Normalized Appearance Stream.} The Normalized Appearance Stream processes 3D coordinates and normalized UV texture through four Omni-scale Modules. Each module incorporates a UV-space adapted Dynamic Graph Convolution (DGC) layer and three parallel branches with varying grouping rates. The network transforms the initial $m \times 6$ input (UV coordinates + RGB) into a $96 \times 512$ feature map, which is then reduced to a 512-dimensional feature vector via global pooling and a fully connected layer. The training utilizes cross-entropy loss for identity classification, with Adam optimization and cosine learning rate scheduling over 1000 epochs. During inference, the final 512-dimensional feature vector serves as a robust representation of normalized appearance across multiple frames.

\vspace{6px}
\hspace{-12px}\textbf{Configuration for Stream 3: Multi-Scale Attention Stream.} The CLIP ViT-L/14 vision encoder is employed to extract multi-scale features, utilizing a Pad-and-Resize technique to uniformly adjust each frame to a resolution of $224 \times 224$, thereby preserving the original body proportions. The resulting feature volume, $\mathbf{G} \in \mathbb{R}^{T \times 257 \times d}$, encapsulates the embedding size of the token, $d$.
Following the methods~\cite{lin2022frozen,ye2024biggait}, the feature maps from the last four layers of the image encoder (i.e., $N=4$) are utilized. Additionally, four Transformer decoder blocks ($M=4$) are applied to further process these features.

\begin{figure}[!htbp]
    \centering
    \includegraphics[width=1\linewidth]{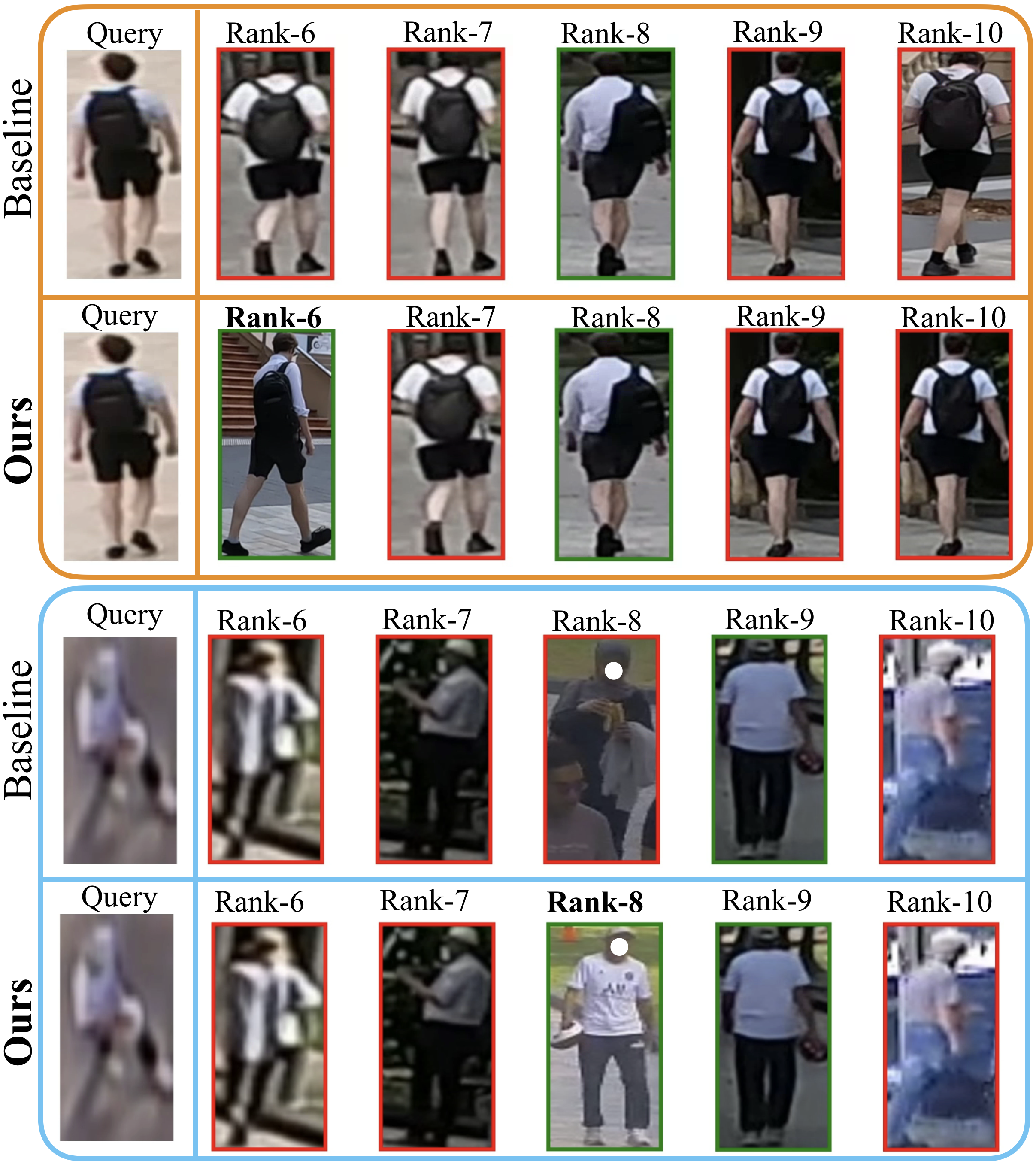}
    \caption{A2G: Comparing baseline vs our method on AG-VPReID. Green/red: correct/incorrect labels. Altitudes: 15m (orange), 120m (blue). First tracklet image is shown. Ranks show improvements in \textbf{bold}. Best in color. }
    \label{fig:aerial_to_ground}
\end{figure}

\section{Visualization}

\begin{figure}[!htbp]
    \centering
    \includegraphics[width=1\linewidth]{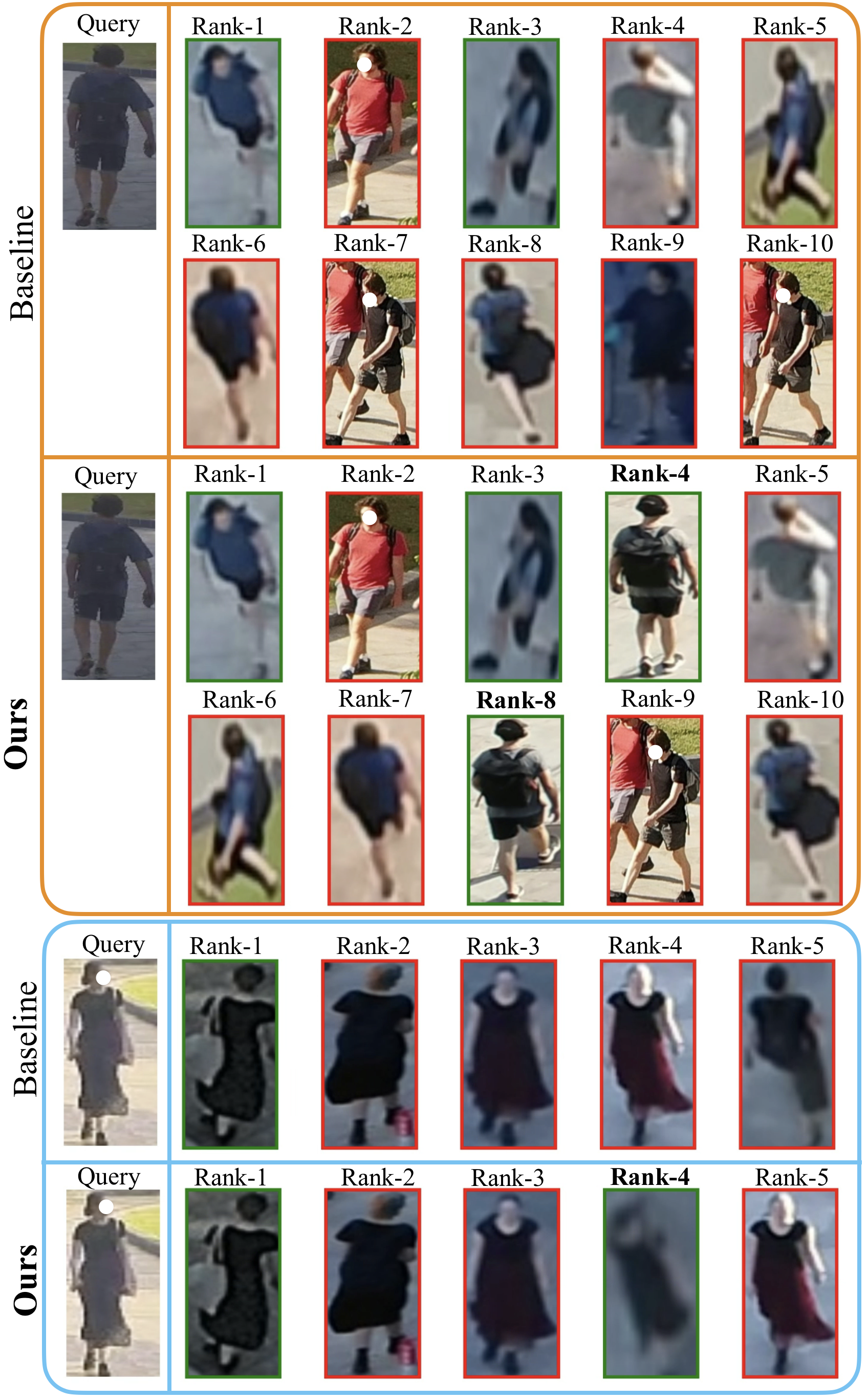}
    \caption{G2A: An extended case of Fig.~\ref{fig:aerial_to_ground}}
    \label{fig:ground_to_aerial}
\end{figure}

To provide a comprehensive analysis of our method's effectiveness compared to the baseline CLIP-based approach~\cite{li2023clip}, we present qualitative results through visualization examples in Fig.~\ref{fig:aerial_to_ground} and Fig.~\ref{fig:ground_to_aerial}. For the aerial-to-ground (A2G) matching scenario shown in Fig.~\ref{fig:aerial_to_ground}, we compare the retrieval results between our approach and the baseline method. Each query image is captured from aerial views at different altitudes (15m and 120m), with corresponding ground-truth matches from ground-level perspectives. The green and red boxes indicate correct and incorrect matches respectively, while improved rank positions are highlighted in bold. These visualizations demonstrate our method's superior ability to handle severe viewpoint variations and maintain reliable person matching across aerial and ground views. Additional examples for ground-to-aerial (G2A) matching presented in Fig.~\ref{fig:ground_to_aerial} further validate the robustness of our approach.

\clearpage

\end{document}